\pdfoutput=1
\documentclass[11pt]{article}

\usepackage[preprint]{acl}

\usepackage{times}
\usepackage{latexsym}
\usepackage{array}
\usepackage{subcaption}
\usepackage{multirow,multicol}
\usepackage[T1]{fontenc}
\usepackage[utf8]{inputenc}
\usepackage{minitoc}
\usepackage[most]{tcolorbox}

\usepackage{microtype}
\usepackage{amsmath}
\usepackage{pifont}
\usepackage{amssymb}
\usepackage{inconsolata}

\usepackage{graphicx}
\usepackage{colortbl}
 \usepackage{booktabs} 
\newcommand{\xmark}{\text{\ding{55}}}%
\usepackage{xspace}
\usepackage{hyperref}
\hypersetup{
    colorlinks,
    linkcolor={black},
    citecolor={black},
    urlcolor={black}
}
\usepackage[capitalize]{cleveref}
\crefname{section}{Sec.}{Section}
\Crefname{section}{Section}{Sections}
\Crefname{table}{Table}{Tables}
\crefname{table}{Table}{Table}

\newcommand{ \ours}[0]{InstructAny2Pix}

\usepackage{tabularx}
\usepackage{amsmath}
\usepackage{array}

\newcolumntype{H}{>{\setbox0=\hbox\bgroup}c<{\egroup}@{}}
\newcolumntype{a}{>{\columncolor{Gray}}c}
\usepackage{color, colortbl}
\newcommand{\ourtitle}{InstructAny2Pix: Image Editing with Multi-Modal prompts.}
\title{ \ourtitle} 
\author{
 \textbf{Shufan Li \textsuperscript{1}},
 \textbf{Harkanwar Singh\textsuperscript{1}},
 \textbf{Aditya Grover\textsuperscript{1}}
\\
\\
 \textsuperscript{1}University of California, Los Angeles
\\
 \small{
   \textbf{Correspondence:} \href{mailto:email@domain}{jacklishufan@cs.ucla.edu}
 }
}

\begin{document}
\maketitle
\begin{abstract} Image editing has made incredible progress in recent years. Early works only supported caption-guided editing, but recently, free-form text instructions and reference images have been incorporated to allow for more flexibility. However, existing methods still struggle with complex editing instructions involving multiple objects or reference images. We present \ours, a novel image editing model that leverages a multi-modal LLM to execute intricate edit instructions. Compared with previous works, \ours~ extends the flexibility of edit instructions in three key ways: First, it can perform complex instructions involving multiple object edits; second, it supports the interleaving of text instructions with multiple reference images; and third, it supports audio and music inputs as part of the edit prompts, unlocking creative applications such as album cover generation and music-inspired merchandise design. To evaluate the effectiveness of \ours, we propose two new benchmark datasets, MM-Inst and Dream-booth++, consisting of human-written, multi-modal prompts. \ours~ outperforms baselines on these two proposed multi-modal benchmarks, as well as on conventional image editing benchmarks such as InstructPix2Pix. \end{abstract}    
\section{Introduction}
\label{sec:intro}

\begin{figure*}[t]
    \centering
    \includegraphics[width=1.0\textwidth]{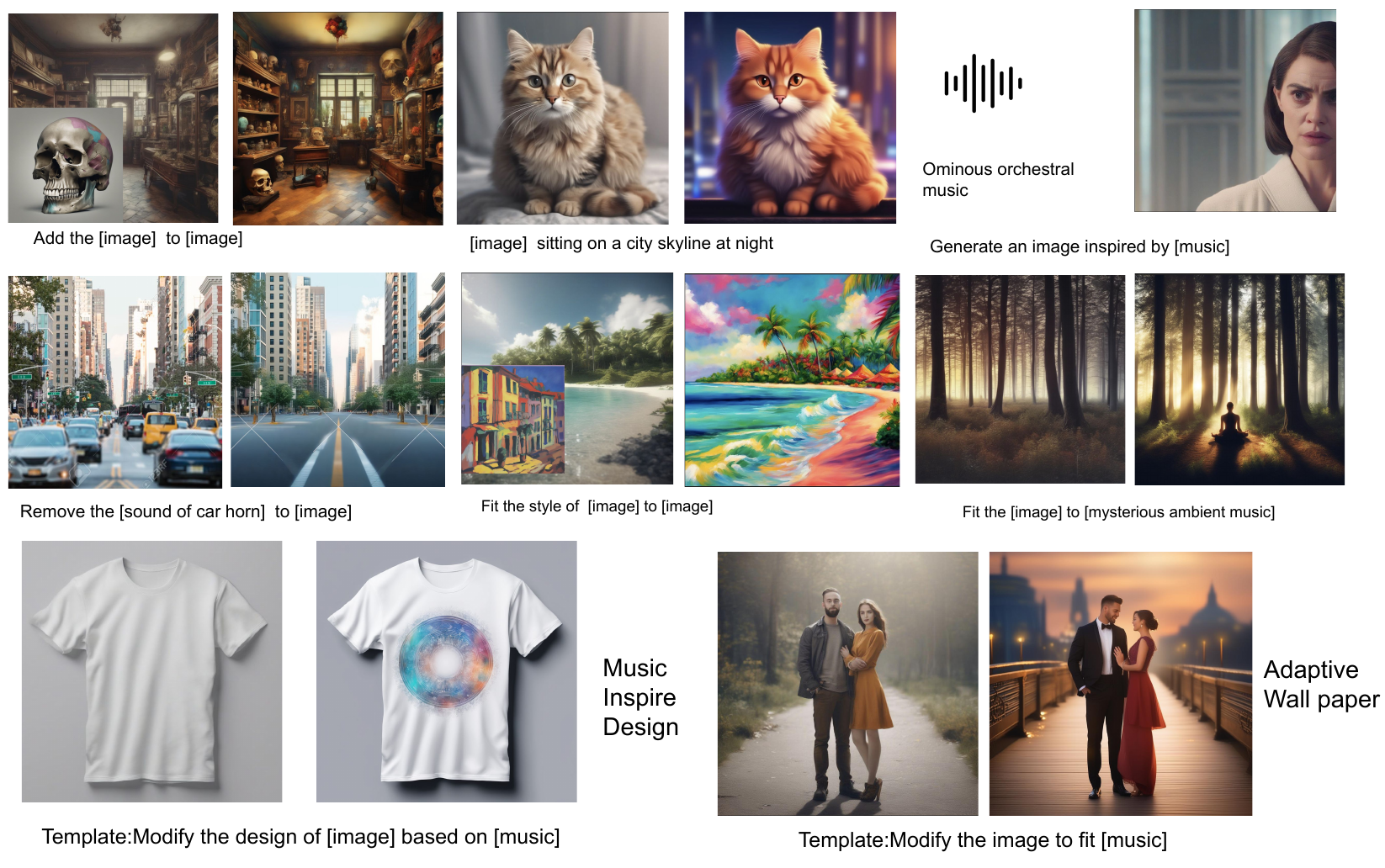}
    \caption{
    Illustration of \ours{}'s ability to flexibly edit an image based on a variety of multi-modal instructions. More examples of audio-guided editing are provided in the supplementary demo video.}
    \label{fig:teaser}
\end{figure*}

The ability to edit an existing image using free-form text instructions vastly expands the usability of image editing models. Compared with early works, such as Prompt2Prompt\cite{hertz2022prompt}, which require caption pairs, instruction-based image editing methods, such as InstructPix2Pix\cite{brooks2023instructpix2pix}, offer users unparalleled flexibility to describe edit instructions in natural language, such as "add a dog." More recent models, such as Kosmos-G\cite{kosmos-g}, additionally accept reference images, allowing users to add a specific dog from the reference image to the scene. Despite these progresses, existing methods still have limited instruction-following capabilities. For text-guided edits, they are limited to simple instructions on which they were trained and cannot generalize to complex instructions involving multiple objects, such as "add a wolf howling under the moon." For image-guided edits, they often struggle to complete complex instructions, such as "replace the cat with [reference image]," while faithfully respecting both the image to edit and the reference image.

To address these limitations, we propose \ours, the first instruction-following image editing system capable of following a wide range of complex, multi-modal, multi-object instructions. Specifically, \ours~not only supports text instructions involving multiple objects, such as "add a wolf howling under the moon" or "add a cat and remove the dog," but it can also optionally accept multiple reference images of the objects (e.g., the wolf and the moon). Furthermore, it works with arbitrary free-form, multi-modal instructions interleaving text, image, and audio, such as "change [image A] to the style of [image B]" or "fit [image] to [music]," while previous multi-modal models only support limited modalities (i.e., image) and very basic instructions (e.g., add, remove).

\ours~greatly enhances the flexibility and usability of image editing models. When creating a scene with multiple objects, instead of writing lengthy descriptions for each object, uploading reference images can be far more efficient. Music or audio inputs, though less obvious, also unlock creative possibilities, such as designing T-shirts based on music or dynamically adapting a background image during live performances. While these tasks could be done through text instructions, they would require designers to first develop specific ideas like "add a circle to the T-shirt" or "change the background to sunset," which demands artistic intuition and experience. Using music as a prompt, however, simplifies this process by reducing the creative burden. Even if the designer isn't explicitly seeking a music-inspired design, experimenting with various tracks and selecting from generated options is quicker than manually drafting multiple design ideas or refining text prompts. Examples of these innovative applications are shown in \cref{fig:teaser}.

Concretely, we build \ours~by combining a multi-modal encoder that "perceives" audiovisual inputs, a large language model that "reasons" about the edit instructions, and a diffusion model that "draws" the edited results. To achieve flexible image editing with multi-modal prompts, we curated a large training dataset of diverse multi-modal editing instructions in three steps. In the first step, we prompt a large language model (LLM) to generate a diverse set of complex, multi-modal edit instructions and captions of intended edit results. Since the LLM cannot generate reference images and audio, we ask it to generate captions of these multi-modal prompts instead. In the second step, we use off-the-shelf text-to-image and text-to-audio models to create reference images and audio from the captions generated in the previous step. In the final step, we employ a pool of caption-based edit methods alongside segmentation and in-painting models to generate edit results using the input images and the captions of intended edit results.

To evaluate \ours~on the proposed tasks, we created two benchmark datasets: MM-Inst and Dreambooth++. Both datasets consist of high-quality, human-written, multi-modal edit instructions. MM-Inst comprises complex multi-modal edit instructions interleaving text, image, and audio. Dreambooth++ specifically focuses on image editing with reference images. Through extensive experiments, \ours~outperforms existing baselines on these two benchmarks. \ours~also achieves competitive performance on simpler instruction datasets, such as InstructPix2Pix, in a zero-shot manner, highlighting \ours's ability to adapt to unseen prompts. After fine-tuning on the InstructPix2Pix dataset, \ours~was able to outperform existing baselines.

\begin{figure*}[t]
    \centering
    \includegraphics[width=0.8\linewidth]{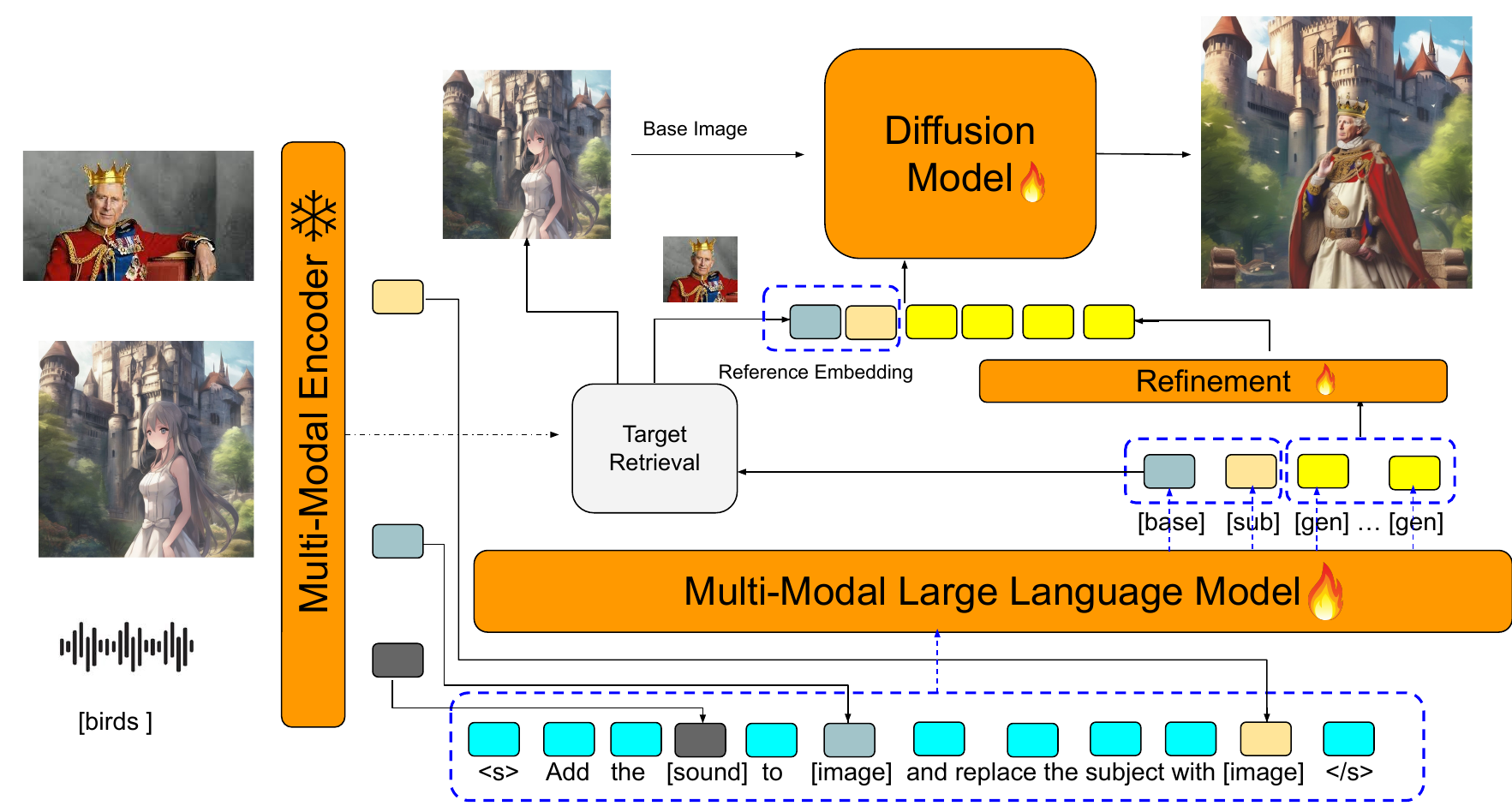}
    \caption{The \ours{} pipeline consists of three building blocks: a multi-modal encoder that "perceives" audiovisual inputs, a large language model that "reasons" about the edit instructions, and a diffusion model that "draws" the edited results. For improved training and generation, we include an additional refinement module to refine the LLM outputs.}
    \label{fig:explainer}
\end{figure*}

\section{Related Works}
\label{sec:related_works}

\subsection{Instruction-Guided Image Editing}
There are numerous image-editing methods based on text-to-image diffusion models \cite{ramesh2022,rombach2021highresolution,podell2023sdxl,kawar2023imagic}. The earliest works required a pair of source and target prompts to perform an edit. Common approaches include DDIM \cite{song2020denoising}, Prompt2Prompt (P2P)~\cite{hertz2022prompt}, Plug-and-Play~\cite{Tumanyan_2023_CVPR}, and Null-text Inversion~\cite{mokady2022null}. These models have very limited flexibility. To achieve good editing results, users must provide long, detailed captions paired in a specific way.

By contrast, instruction-guided image editing methods only require vague text instructions, such as "add fireworks." InstructPix2Pix \cite{brooks2023instructpix2pix} first achieved this by curating a large machine-generated image-editing dataset using P2P and then directly fine-tuning a diffusion model end-to-end on this dataset. MagicBrush \cite{Zhang2023MagicBrush} curated a higher-quality human-annotated dataset by requesting humans to perform editing operations using tools such as Photoshop. MGIE \cite{fu2023guiding} utilizes a multi-modal large language model to process editing instructions and input images. While it achieves better results than pure diffusion-based methods like InstructPix2Pix and MagicBrush, it still operates only with a single source image and text-only instructions. 

Unlike previous works in this area, our work extends the edit instructions to multi-modal, multi-object instructions, greatly enhancing the flexibility of image editing models.

\subsection{Multi-Modal Conditioned Generation}
Parallel to these image editing methods, there have been previous attempts to achieve image generation with multi-modal conditioning using multi-modal language models. BLIP-Diffusion \cite{li2023blip2} incorporates BLIP \cite{li2023blip} as a multi-modal encoder that generates subject embeddings for the diffusion model. Using this approach, it can generate images following text prompts and reference images. Kosmos-G \cite{kosmos-g} directly aligns the representation space of multi-modal language models with that of a diffusion model. Kosmos-G allows image generation based on multiple reference images. However, since these works focus on generation rather than instruction-based image editing, they support neither the removal and replacement of objects nor other free-form instructions. They also cannot faithfully respect the spatial structure of input images. 

Audio-guided image generation is a relatively uncharted area. AAI \cite{yang2023align} achieves sound-guided generation by aligning audio representations to reference images. This method is very limited in that it requires retrieving 3-5 reference images and performing gradient descent optimization steps for each audio input at inference time.

Unlike previous works in this area, our work is the first to support interleaved audiovisual inputs and free-form image editing instructions.

\section{Methods}
\label{sec:methods_all}
\begin{figure*}[t]
    \centering
    \includegraphics[width=1.0\linewidth]{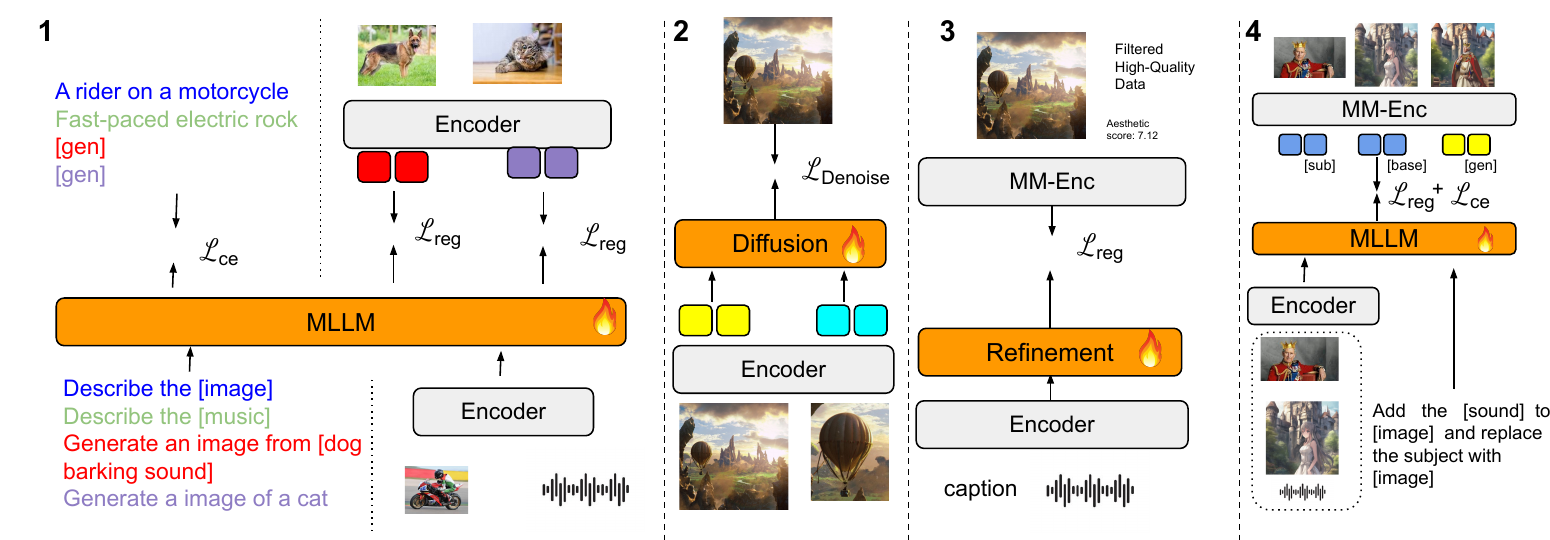}
    \caption{Training pipeline of \ours{} consists of four steps. 1. Pretraining of Multi-Modal LLM with text-to-x and x-to-image tasks. 2.  Pretraining of Diffusion Decoder  3. Pretraining of Refinement Module. 4.Instruction Finetuning }
    \label{fig:training_pipeline}
\end{figure*}

\subsection{Model Architecture}
The architecture of \ours~ is illustrated in Figure \ref{fig:explainer}. It consists of a multi-modal encoder that maps multi-modal inputs to a unified latent space, a multi-modal LLM that generates a set of edit tokens autoregressively, and a diffusion image decoder that generates the edit result conditioned on the input image and edit tokens. We initialize the multi-modal encoder with Imagebind\cite{girdhar2023imagebind}, the LLM with Vicuna-7B\cite{vicuna} and the diffusion image decoder with SDXL\cite{podell2023sdxl}. 

The input of \ours~ consists of a multi-modal instruction $(T,I,A)$. It includes a text instruction $T$, a set of images $I$ and a set of audio pieces $A$. $I$ contains the input image and optional additional reference images. It is always non-empty. $A$ contains optionally referece audio pieces. It can be an empty set. 

We first leverage the multi-modal encoder $Enc$ to encode reference images and audios into embeddings $E_I=Enc(I),E_A=Enc(A)$. We use the token embedding layer $Emb_{LM}$ of the LLM to obtain the text embedding $E_T=Emb_{LM}(T)$. We then interleave the text, image and audio embeddings into a sequence of instruction embeddings $E_{inst}$.  These embeddings are then passed to the large language model, which generates a sequence of \textit{discrete} control tokens $C$ autoregressively. For each token, we also extract the corresponding \textit{continuous} hidden state from the last transformer block to form a sequence of control embeddings $E_c=(E_{base},E_{sub},E_{gen})$.

Each control token belongs to one of the following type: [base], [gen], [sub]. The embedding of the [base] token is used to retrieve the input image. The embedding of [sub] tokens is used to retrieve reference images of relevant subjects in the prompt. Typically there is only one [base] token, but there can be more than one [sub] tokens to account for multiple reference images. [sub] token will only be generated if the referenced object should appear in the final image. For example, in instruction "replace [image of object A] with [image of object B]", only the [sub] token corresponding to [image of object B] will be generated.

The [gen] embedding $E_{gen}$ are further processed by a refinement module. The output of the refinement module $E_{refined}$ is used to condition the diffusion image decoder $Dec$ alongside retrieved images to create the final edit result $X_{out}$.  The whole process can be summarized in the following formula.  

\begin{align}
    E_A = Enc(&A), E_I = Enc(I)  \\
    E_T &= Emb_{LM}(T)  \\
    E_{inst} = Int&erleave(E_T,E_A,E_I) \\
    (E_{base},E_{sub},&E_{gen}) = LLM(E_{inst}) \\
    E_{refined} &= Refinement(E_{gen}) \\
    X_{out} &= Dec(E_{refined},\\
    &  \quad Retrieve(E_{base},E_{sub}))
    \label{eq:a2p}
\end{align}

To account for mismatches in the dimension of embeddings, we add MLP projectors where needed. We provide additional details in \cref{sec:appendix_model_arch}.

\subsection{(Continued) Pretraining}
\subsubsection{Diffusion Image Decoder}
The diffusion decoder was initialized with SDXL, which generates images based on CLIP text embeddings. At inference time, we want the decoder to generate images based on $E_{gen}$, which encodes some global semantics, and reference images retrieved by $E_{sub}$, which contain some objects that should appear in the edit result. To achieve this end, we repurposed SDXL to generate images based on the CLIP image embedding of the whole image $E_{global}$, and the CLIP image embeddings of objects in the image $E_{obj}$. When performing an image edit, we use $E_{gen}$ generated by LLM as $E_{global}$, and the CLIP image embeddings of retrieved reference images as $E_{local}$. 

To obtain object-level encodings $E_{local}$, we employ an object detector to find the bounding boxes of objects in the images. We crop the image using the bounding boxes, and use the CLIP embedding of the cropped image as $E_{local}$. To prevent biases and limitations of the object detector  (e.g. the detector cannot detect certain classes), we also incorporate additional bounding boxes sampled randomly from a uniform distribution. We randomly drop $E_{global}$ and $E_{obj}$ independently during the training to prevent the model from over-relaying on one of the embeddings while ignoring the other.

We use a subset of LAION-Aesthetics-V2 \cite{schuhmann2022laion} dataset with 4M images for this task. We only use the images in the dataset, ignoring the text captions. 

\subsubsection{Multi-Modal LLM}
We initialize the LLM with Vicuna-7B \cite{vicuna}. Since Vicuna was only trained for language modeling, we continue to pretrain it on multi-modal data consisting of images, texts, and audio. We use text-image pairs from LAION-Aesthetics-V2, text-audio pairs from Audioset\cite{gemmeke2017audio}, LP-MusicCaps \cite{doh2023lp}, Audiocaps\cite{audiocaps}, and audio-image pairs from SoundNet\cite{aytar2016soundnet}, VGGSound\cite{chen2020vggsound}.  We compose multi-modal prompts for 4 tasks as our pretraining objective: image captioning, audio captioning, text-to-image generation, and audio-to-image generation. Examples of these prompts are "describe the [image]", "generate an image of a cute dog running in a garden", and "generate an image based on [audio]". For captioning task, we apply autoregressive next-token prediction loss. For generation tasks, we set the target token to "[gen]" and apply next-token prediction loss. Additionally, we also extract the embedding $E_{gen}$ and apply L2 regression loss in the embedding space. For example, if the prompt is "generate an image based on [audio]", we apply the next-token classification loss between the output logits and the target sequence "[gen]", and apply the L2 regression loss between the output embedding and the visual embedding of the target image.

\subsubsection{Refinement Model}

After pretraining the Diffusion Image Decoder and Multi-Modal LLM independently, we observe that directly use the output embedding $E_{gen}$ of LLM as the conditioning embedding $E_{global}$ of the image decoder leads to very low-fidelity image outputs. This occurs because most image-audio pairs in the training data come from low-fidelity YouTube videos (LAION Aesthetic Score < 4.5), while typical text-to-image training schemes use datasets with high-fidelity images (LAION Aesthetic Score > 5.0). To mitigate the effect of low quality data, we incorporate a refinement module, which is a transformer that learns to improve the image quality in the embedding space. 

Concretely, the refinement module takes an image embedding $E$ and a target aesthetic score $s$, and generates $E_{refined}$, which is the embedding of an image with the same semantics but achieves the target fidelity. Since pairs of two images that are identical in semantics but different in fidelity are hard to find, we make use of existing image-text and image-audio pairs to generate $(E,E_{refined},s)$ triples for training. Given an image-text or image-audio pair, we use the visual embedding of the image as $E_{refined}$, and the aesthetic score of the image as $s$. We use the corresponding text or audio embedding as $E$.  While $E$ is not an actual image embedding, it is still an embedding representing the same semantics, since the latent space of different modalities are aligned through the multi-modal encoder.

At inference, we use the LLM output embedding $E_{gen}$ as $E$ and set the target score $s$ to a high number (e.g. 6.0).  This process is shown in \cref{fig:training_pipeline}.

\subsection{Instruction Fine-tuning}

\subsubsection{Data}
\label{sec:main-mm-inst-data}
To train \ours~ for image editing tasks, we curated a diverse dataset of 500k instructions and corresponding image pairs, called MM-Inst. The dataset generation pipeline consists of three steps: 
text instruction generation, reference multi-modal inputs generation, and input-output image pair generation. 

In the first step, we prompt a Large Language Model (LLAMA2 \cite{touvron2023llama}) to generate creative instructions using 36 manually written examples. Since the LLM cannot generate reference images, audio and music, we ask the LLM to generate the captions of multi-modal references instead. To further increase the diversity of instructions, we prompt the language model to generate instructions based on ground-truth music captions from LP-MusicCaps and AudioCaps, as well as ground-truth image captions from LAION.  We provide further details in \cref{sec:appendix_training_dataset}

In the second step, we curate the corresponding reference images and audios using the captions created in the first step. If a caption is a ground-truth image, audio or music caption, we directly fetch the corresponding media. If the caption is generated, we use SDXL\cite{podell2023sdxl} and AudioLDM2\cite{liu2023audioldm2} to generate images and audio respectively.

In the last step, we curate pairs of input images and edit results using a combination of six methods: 

1) Edit image using captions and Prompt2Prompt \cite{hertz2022prompt} 2) Edit image using captions and Plug-and-Play \cite{Tumanyan_2023_CVPR} 3) Use DDIM\cite{song2020denoising} inversion on the source image and generate a new image using target prompt with inversed latent. 
4) For object removal, we use an open-vocabulary object detector (GroundingDINO \cite{liu2023grounding}) to locate the object and perform inpainting in the area of the removed object. 5) For object addition, we first generate an image using target prompt as the target image. Then we use the detector to localize the added object and remove it through inpainting. The resulting image is used as the source image. 6) For object replacement, we first follow the removal procedure. Then we perform inpainting in the area of removed objects using the replacement object as prompts.

When the detector fails to localize the object or yields low confidence scores, we fall back to caption-based methods. Following InstructPix2Pix \cite{brooks2023instructpix2pix}  we filter the results using CLIP scores. We additionally filter the results using the LAION aesthetic predictor and remove low quality images.  We also provide additional details in \cref{sec:appendix_training_dataset}.
% show this pipeline in \cref{fig:dataset}

\subsubsection{Training}
\subsection{Instruction Guided Finetuning}
We fine-tune the LLM using the same objective as the continued pretraining phase, which consists of a next-token prediction loss on the output logits and a regression loss on the output embeddings. Unlike the continued pretraining phase which only includes simple generation or captioning tasks, we use interleaved multi-modal instructions from MM-Inst dataset as the input. We formulate the target output sequence as "[base] [sub] .. [sub] [gen] .. [gen]"  where each [sub] corresponds to a reference image. We only add a "[sub]" token if the referenced object should appear in the desired output. For example, in the following instructions "add [a]", "remove [b]", "replace [c] with [d]", only [a] and [d] will have corresponding [sub] tokens in the target sequence. L2 regression loss is applied between the [base] embedding and the visual embedding of the source image, between [sub] embeddings and the visual embeddings of reference images, and between the [gen] embedding and visual embeddings of the edit results. The diffusion model is not directly used in this process. 

\section{Evaluation}
\label{sec:dataset_building}

% The diffusion model is conditioned on the embedding generated by the multi-modal encoder and trained using the standard diffusion loss \cite{ho2020denoising}:
% \begin{equation}
% \mathcal{L}_{\text{diff}} = \mathbb{E}_{t} \left[ \lVert \epsilon_{t} - F_{\text{diffusion}}(z_t,F_{\text{enc}}(X)) \rVert_2^2 \right]
% \label{eq:diff_loss}
% \end{equation}
% where $z_t$ is a noised version of the latent in the diffusion process and $\epsilon_t$ is the noise at time $t$.
% We use 2M samples of high-quality images from LAION-Aesthetic-V2 datasets \cite{schuhmann2022laion} and SDXL \cite{podell2023sdxl} as our base model.

% \subsubsection{Prior Alignment}

% To train our refinement prior, we use 2M text-image pairs from LAION-Aesthetic-V2 datasets, which consist of images with high visual quality and corresponding captions. Because the vanilla captions are of low quality, we augment the captions by recaptioning the images using BLIP2~\cite{li2023blip}. We also make use of 2M audio-visual pairs sampled from VGG-sound~\cite{chen2020vggsound}, AudioSet~\cite{gemmeke2017audio}, and SoundNet~\cite{aytar2016soundnet}. This allows the prior model to learn a broad range of textual-visual and audio-visual correspondences. We also train a version of our prior in a bidirectional way, which means our prior can also be used to predict audio tokens given images. This version is used in our data generation pipeline. The details will be discussed in the following section.

% \subsubsection{Instruction Tuning}

    \subsection{Evaluation Dataset}
\label{sec:eval_data}
Image editing following multi-modal instruction is a novel task with no existing benchmarks. To fairly evaluate \ours's  performance in real-world settings, we curated 1500 manually written multi-modal instructions. Unlike previous works which perform evaluations on samples from the same distribution as their training data, our evaluation benchmark caters for diverse real-world usecases. We call this dataset MM-Inst-Test. We provide further details in \cref{sec:appendis_evaluation_dataset}.

For image guided generation, Dreambooth\cite{ruiz2023dreambooth} is a commonly-used benchmark. However we find it inadequate to provide a holistic evaluation of multi-modal generative models. Firstly, it only contains two classes of live objects (cats and dogs), which accounts for 9 out of 30 subjects in the dataset. Its diversity is limited. Secondly, its task only involves changing the background of a single subject. This setup cannot evaluate a model's capability of making use of multiple image inputs. To address this gap, we propose Dreambooth++, a Dreambooth-like dataset with more diverse prompts. It consists of 30 subject images which are evenly distributed across humans, animals, small objects, and large structures. We also include 30 diverse background images with corresponding prompts. In total, there are 900 generation tasks. We evaluate models on this dataset using two protocols:  single-image and multi-image. The single-image setup is similar to Dreambooth, which requires generating a given subject under different context (background) prompts. The multi-image task requires generating a new image by combining a subject image and a background image. We provide more details in \cref{sec:appendis_evaluation_dataset}

Additionally, for completeness and fair comparison with existing models, we evaluate our model on 1000 samples from InstructPix2Pix dataset. We report both the zero-shot results and fine-tuned results.

\begin{table*}[t]

    \centering
    \caption{Multi-modal image editing on MM-Inst-Test dataset and Text-based Image editing on InstructP2P dataset. (I+T+A) and (T) refers to using multi-modal instruction and text-only instruction respectively. The best number is \textbf{bolded}. We report both zero-shot and fine-tuned performance on InstructP2P. All baseline methods are trained on  InstructP2P. Win rate represents the percentage of human responses that prefer \ours~ over baseline methods. We use the zero-shot results (Row 2) for human eval on InstructP2P dataset.}
    \begin{tabular}{c cccHc cccHc}
        \toprule
         & \multicolumn{5}{c}{MM-Inst}  &  \multicolumn{5}{c}{InstructP2P}\\
         \cmidrule(r){2-6} \cmidrule(r){7-11}
         & CLIP$_{dir}$ & CLIP$_{im}$ & CLIP$_{out}$ & DINO. & Win.  & CLIP$_{dir}$ & CLIP$_{im}$ & CLIP$_{out}$ & Zero-shot & Win. \\
         \hline
     Ours(I+T+A)    & .099  & .816 & .260 & \textbf{-} & \textbf{-} \\
      \hline
          Ours(T,Zero-Shot)    & \textbf{.095}  & {.856} & \textbf{.270} & \textbf{-} & - & {.147} & {.808} & {.312} & \checkmark & - \\
                    Ours(T,Finetuned)    & -  & - & - & \textbf{-} & - & \textbf{.182} & \textbf{.873} & \textbf{.323} & \checkmark & - \\
      InstructP2P      & {.091} & .824 & {.243} & - & .712 & .145 & .742 & .241 & \xmark & .646 \\
      MagicBrush & .084 & .807 & .199 & - & .707 & {.165} & .760 & .250 & \xmark & .698\\
    InstructDiff. & .066 & \textbf{.940} & .193 & - & .746 &.126 & {.857} & .301 & \xmark & .631 \\
  %   \hline
  % Ours(T+I)    &\textbf{.215}  & .759 & \textbf{.271} & .51/.58 & - \\
  % Blip-Diff.    & .132  & \textbf{.790} & .214 & .49/.60 & .790 \\          
  %  Kosmos-G    & .198  & .656 & .270 & .60/.33 & .862 \\
    \bottomrule
    \end{tabular}
    \label{tab:exp1}
\end{table*}

\begin{table*}[t]
    \centering
    \caption{Image conditioned generation on DreamBench++ dataset. $C_{dir},C_{im},C_{out}$ is abbreviated form of CLIP$_{dir}$, CLIP$_{im}$ and CLIP$_{out}$. For multi-image setup, numbers are reported in DINO$_{ref}$/DINO$_{sub}$ format. The best number is \textbf{bolded}. }
    \begin{tabular}{c cccHc ccccc}
    \toprule
         & \multicolumn{5}{c}{Single-Image} &  \multicolumn{5}{c}{Multi-Image}\\
         \cmidrule(r){2-6} \cmidrule(r){7-11}
         & C$_{dir}$ & C$_{im}$ & C$_{out}$ & DINO. & DINO  & C$_{dir}$ & C$_{im}$ & C$_{out}$ & DINO & Recall \\
         \hline
         Ours(T+I)    & \textbf{.147}  & {.810} & \textbf{.260} & \textbf{-} & \textbf{.688} & {.154} & \textbf{.789} & \textbf{.309} & \textbf{.625}/{.471} & \textbf{.841}\\
         BLIP-Diffusion   & .089 & .779 & .231 & - & .660 & .091 & .701 & {.292} & {.526}/.422 & .693 \\
         Kosmos-G & {.126} & \textbf{.843} & {.251} & - & .683 & \textbf{.166} & {.740} & .286 & .485/\textbf{.476} & {.812}\\
    \bottomrule
    \end{tabular}
    \label{tab:exp2}
\end{table*}
% \begin{figure}[t]
%     \centering
%     \includegraphics[width=350px]{assets/image-base.pdf}
%     \caption{Qualitative comparison with Image-Based Methods on Dream Bench++ dataset. }
%     \label{fig:qualitative_blip}
% \end{figure}

% \begin{figure}[t]
%     \centering
%     \includegraphics[width=350px]{assets/figure2.pdf}
%     \caption{We compare against T2I \cite{mou2023t2i} on image variations. We use an official implementation of T2I based on SDXL. We select the examples from demos on the official homepage. We do not introduce novel styles. In particular, ``anime," ``digital art," and ``3D rendering" are provided style choices on the HuggingFace demo. We demonstrate that taking image inputs can improve the quality of editing results, highlighting the benefits of our setup.}
%     \label{fig:t2i}
% \end{figure}

\begin{figure}[t]
    \centering
    \includegraphics[width=1.0\linewidth]{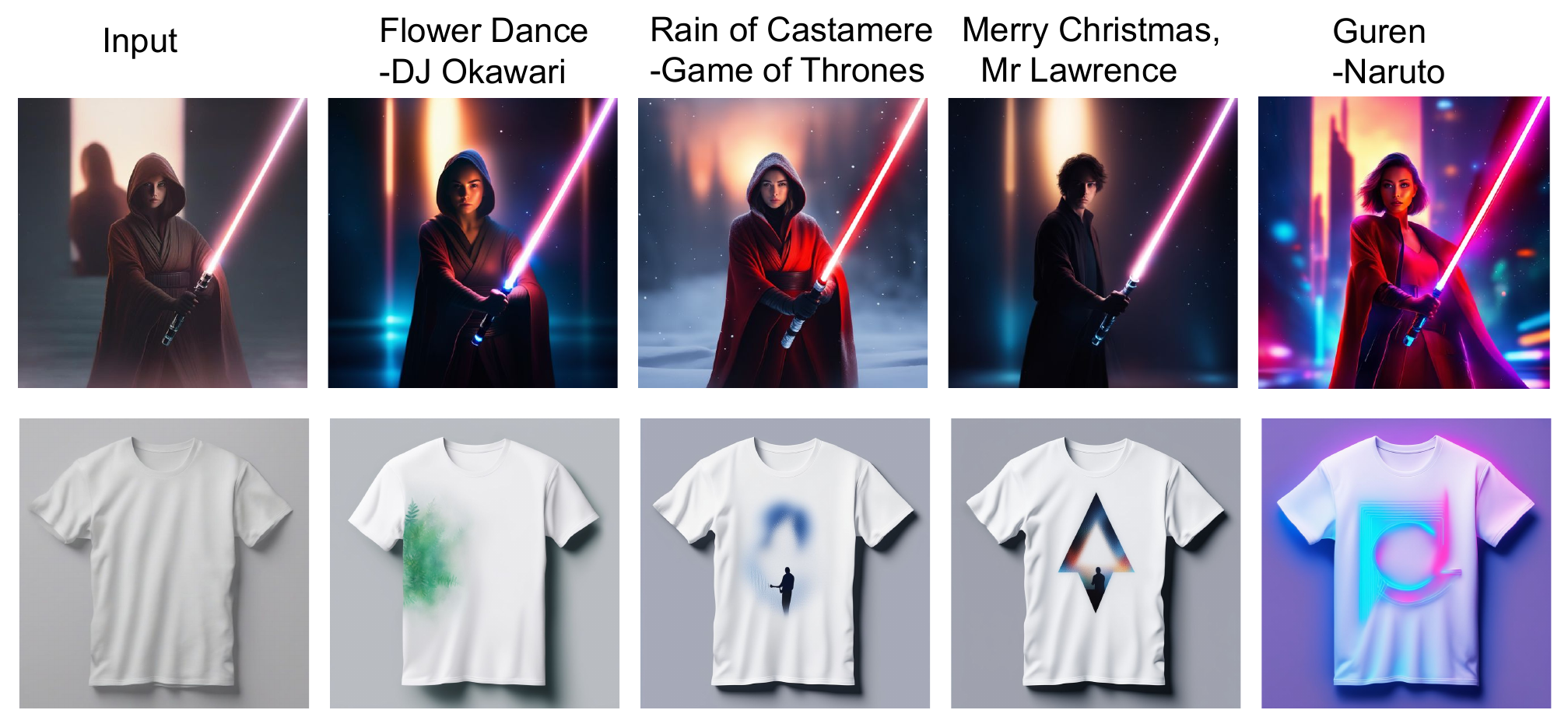}
    \caption{\textbf{Music-Gudied Image Variation}: Music uniquely conveys emotions that are hard to describe using other modalities such as language. We show qualitative results of music guided image variation and music inspired design. \ours~ is able to understand a diverse set of emotions embedded in music and generate creative designs and edits. We include these examples with audio in our supplementary video.}
    \label{fig:music}
\end{figure}

\begin{figure}[t]
    \centering
    \includegraphics[width=1.0\linewidth]{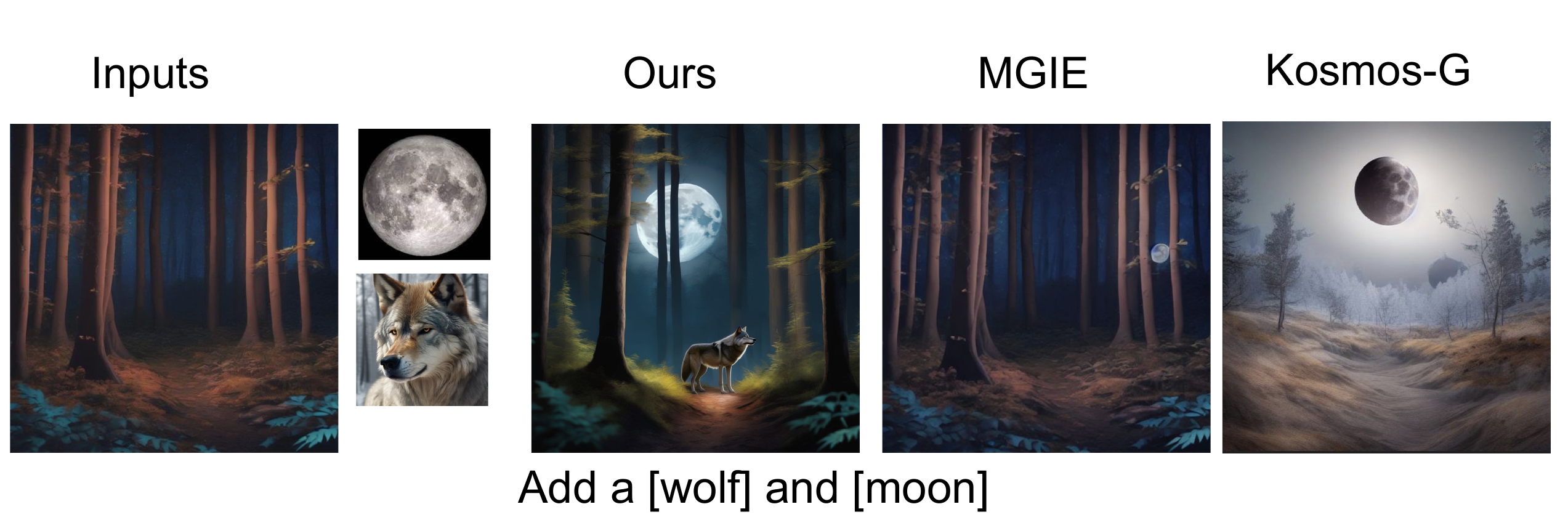}
    \caption{\textbf{Editing with Multi-Object Instructions.} Compared with previous text-based method (MGIE) and image-based method (Kosmos-G), \ours~ uniquely accomplish complex edit tasks. }
    \label{fig:multiturn}
\end{figure}

% \begin{figure}[h]
%     \centering
%     \includegraphics[width=300px]{assets/figure7.pdf}
%     \caption{We provide human evaluation image-to-sound and text-to-image capabilities. For image to audio generation, we leveraged AudioLDM2 \cite{liu2023audioldm} to decode audio embedding to sound. The human reviewers are asked to decide which model is the most preferable among \ours, CoDi \cite{tang2023any}, and NextGPT \cite{wu2023next}. We use this evaluation only to validate our data generation pipeline as we use the prior to generate audio and image embeddings.}
%     \label{fig:codi}
% \end{figure}

% We qualitatively and quantitatively evaluate our results on a wide range of challenging instructions with varying combinations of multi-modal inputs.

\subsection{Instruction guided Image Editing}
We evaluate the capability of \ours~ on MM-Inst-Test dataset described in \cref{sec:eval_data}. Because no previous methods can perform such a task, we selected  text-only instruction models as our baseline. For fair comparison, we convert all multi-modal instructions to text-only instructions using the captions of referenced audio and image. For \ours, we report the performance of using multi-modal prompts and the performance of using text-only prompts. We also report results on 1000 randomly selected images from InstructPix2Pix dataset. We report quantitative metrics in \cref{tab:exp1}. CLIP$_{\text{dir}}$ measures the agreement between changes to captions and changes to images, CLIP$_{\text{im}}$ measures the similarity between the source and targeted images. CLIP$_{\text{out}}$ measures the similarity between edited images and targeted captions. 

We also conducted human evaluations on both dataset and report the win rate. Human evaluators are asked to pick a preferred edit output in a one-to-one comparison between \ours~ and each baseline method. For a fair comparison, we use the text-only version of our method. 

The results are shown in \cref{tab:exp1}. \ours~show decisive advantages in human preference and strong performances in quantitative metrics. We compare with baseline methods InstructPix2Pix\cite{brooks2023instructpix2pix}, Magicbrush \cite{Zhang2023MagicBrush} and InstructDiffusion \cite{Geng23instructdiff}. Notably, we achieve competitive performance on InstructPix2Pix dataset without ever training on such dataset. This result showcases the superiority of our data generation pipeline. It incorporates a diverse range of instructions and enables our model to generalize to unseen instruction patterns. We observe that \ours~ has slightly higher $\text{CLIP}_{\text{im}}$ and $\text{CLIP}_{\text{out}}$ when using only text instructions. This may reflect the fact that multi-modal image editing process are affected by multiple reference images, rather than just an input image and a text instruction.

\subsection{Image Conditioned Generation}
We evaluate \ours~ on DreamBench++ dataset described in \cref{sec:eval_data}. We conduct both single-image and multi-image evaluation and compare results with BLIP-Diffusion \cite{li2023blip2} and Kosmos-G \cite{kosmos-g}. In addition to metrics reported in the previous section, we additionally report DINO scores, which measures the image similarity. For multi-image benchmark, evaluating the DINO similarity of the entire image does not make sense, as the subject is added to the scene may not necessarily occupy the entire image. To address this, we use a segmentation model \cite{liu2023grounding} to segment the subject in generated images. We crop the image according to the bounding box of the object. We report the DINO similarity between the cropped image and the reference subject image as DINO$_{sub}$, and the DINO similarity between the whole result and the background input image as DINO$_{ref}$. We provide quantitative metrics in \cref{tab:exp2} and more qualitative comparison in \cref{sec:appendix-baselines}. \ours~ shows a clear advantage in generation quality and image consistency. 

\subsection{Discussions}

\begin{figure}[h]
    \centering
    \includegraphics[width=0.9\linewidth]{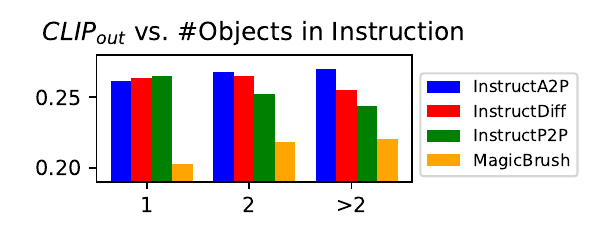}
\caption{\textbf{$\text{CLIP}_{\text{out}}$ with respect to number of objects on MM-Inst-Test dataset.}}
    \label{fig:multiturn-data}
\end{figure}

\textbf{Does \ours~ outperforms baselines on complex, multi-object instructions?} Unlike previous works, \ours~ can perform complex editing operations involving multiple multi-modal inputs. In figure \cref{fig:multiturn}, we provide visual examples of \ours~ performing complex instructions where existing models fail.  We also visualize the performance gap on single-object and multi-object prompts in \cref{fig:multiturn-data}. \ours~ exhibits a larger lead in multi-object prompts. While the numerical performance on instructions with only one object is similar, we still observe qualitative differences. Additional analysis is provided in \cref{sec:appendix-baselines}.

\textbf{Does the audio/music capabilities of \ours~ enable useful real-world applications?} \ours~ uniquely enables music guided image editing, which can be quite useful in cases like music-inspired designs and image variations. We show some examples in, \cref{fig:music}. We recognize that connecting music to visual elements (e.g. smooth music to peaceful scenes) can be a subjective process. We provide further discussion in the appendix \cref{sec:appendix-music}.  We also include an audible demonstration video in the supplementary.
%\cref{fig:multiturn}

% \subsection{Music Guided Generation} \ours uniquely enables music guided image editing, which can be quite useful in cases like automatic album generation using portraits of artists. To evaluate this potential, we perform music-guided image variation where we prompt the model with the instruction "modify [image] to convey the feeling of [music]". We find the model can successfully adapt an image to different music in subtle ways hard to be described in other modalities. 

\section{Conclusion}
In summary, we propose \ours, a flexible system for  editing images based on multi-modal, multi-object instructions. Compared with previous works, it uniquely supports complex multi-object, multi-modal instructions. It also unlocks creative new use cases such as multi-image synthesis and music inspired designs. We proposed two novel benchmarks: MM-Inst-Test and Dreambooth++ for image editing with multi-modal prompts and multi-image synthesis respectively.  \ours~ outperforms existing baselines on these benchmarks, while also achieving competitive performance on conventional image-editing benchmarks with only text instructions.
% incorporating LLMs into image-editing systems.

\section{Limitations}
\label{sec:appendix_limitations}
\subsection{Biases}
Our model makes use of a pretrained diffusion model \cite{podell2023sdxl} and a pretrained LLM \cite{vicuna}. Hence, it may inherit biases from the training process of these models. For example, the SDXL is known to have some biases towards certain skin color \cite{esposito2023mitigating}. Our system will inherit these biases. 

\subsection{Style of Output Images}
Our model tends to bias towards artistic/painting outputs instead of photorealistic ones. This is caused by multiple factors: First, the LAION-Aesthetic-3M \cite{schuhmann2022laion} dataset used to pretrain the diffusion model contains a lot of art and paintings. Additionally, the LAION Aesthetic score used to condition the refinement model is biased towards high saturation and artistic outputs. Lastly, we use SDXL to generate images for the MM-Inst dataset based on captions. Without explicit style keywords in prompts, we find that SDXL generations are biased towards artistic outputs as well. We will try addressing this limitation by exploring alternative ways of curating a high-quality dataset and explicitly adding diverse style prompts in the generation process.

\subsection{Types of Supported Edits}
In this work, we explore mostly object-level edits and global edits that change the semantics of images, such as adding and removing objects, changing backgrounds, changing the image style, and changing the overall atmosphere of the scene.  \ours~ does not currently support Photoshop-style edits such as increase the image brightness, zoom in on objects. Users may choose to fine-tune \ours on relevant datasets. We left that for future exploration.

\bibliography{egbib}

\appendix

% \label{sec:appendix}

\setcounter{figure}{0} % Reset figure counter
\setcounter{table}{0}  % Reset table counter
\renewcommand{\thefigure}{A.\arabic{figure}} % Set figure numbering
\renewcommand{\thetable}{A.\arabic{table}}   % Set table numbering
\sloppy
{

        {\centering
        % \Large
        \textbf{\ourtitle}\\
        \vspace{0.5em}Appendix \\
        \vspace{1.0em}}
   }

% \addcontentsline{toc}{section}{Appendix} % Add the appendix text to the document TOC
% \part{} % Start the appendix part
% \parttoc % Insert the appendix TOC
\section{Table of Contents of Supplementary Material}

In the supplementary software submission, we provide the code of \ours. The prompt used to generate the training data is also provided with the code. In the supplementary data submission, we provide the full evaluation dataset, including MM-Inst-Test and Dreambooth++. We also include a demo video with various examples shown in the paper, accompanied by audible music and sound. 

\begin{figure*}[t!]
    \centering
    \includegraphics[width=1.0\linewidth]{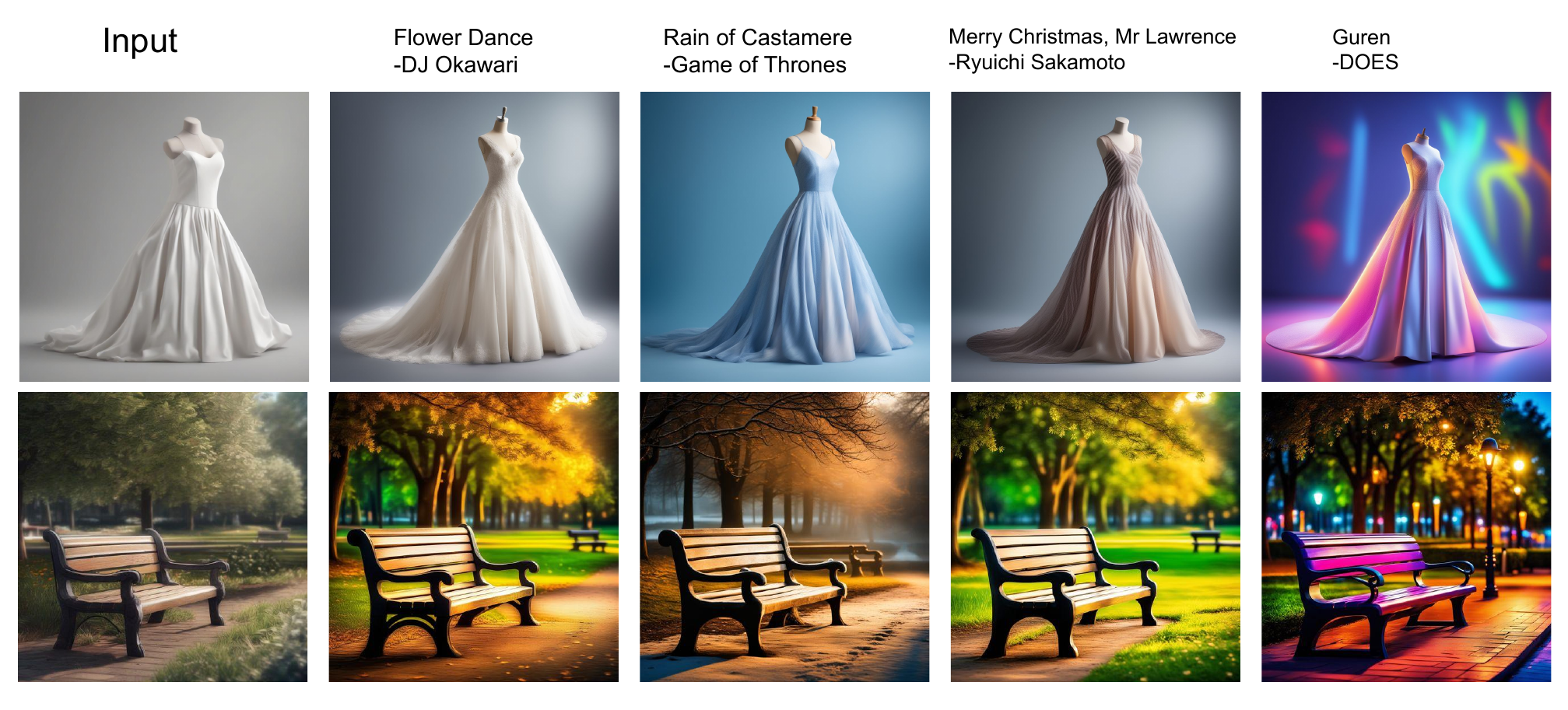}
    \caption{Additional qualitative results of music inspired designs and image variations. \ours~ was able to make diverse image edits given a music prompt. We include these examples with audio in our supplementary video.}
    \label{fig:musicv3}
\end{figure*}

\begin{figure*}[h!!]
    \centering
    \includegraphics[width=1.0\linewidth]{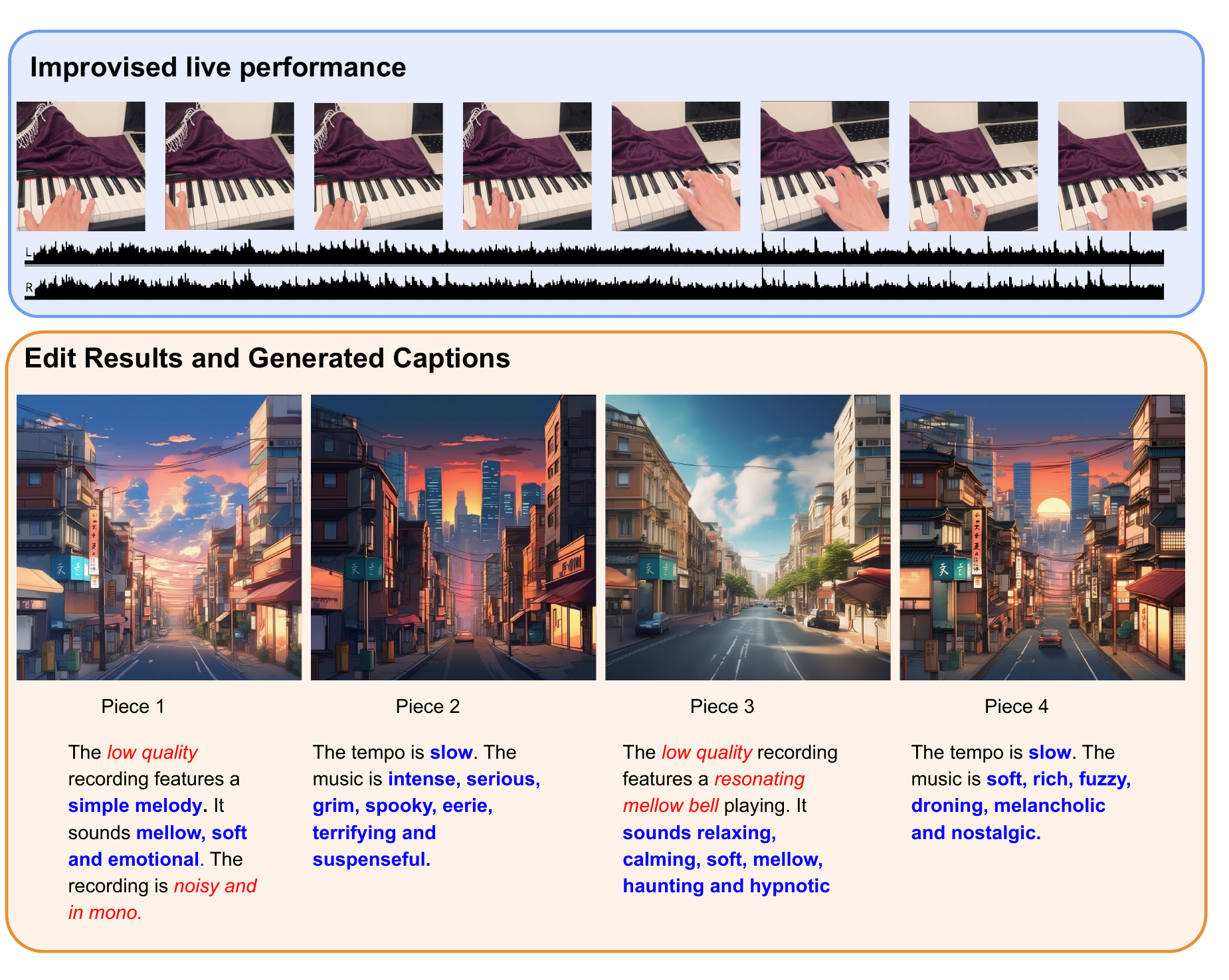}
    \caption{Qualitative results of live performance visuals. We use \ours~ to generate a set of visuals corresponding to an improvised performance involving four pieces. We also prompt the model to generate the captions of each music to understand its reasoning process. We mark attributes that are reasonable interpretations of the music with blue, and unreasonable interpretations with red. To maximize creativity, we use a low CFG for this task for better image diversity. }
    \label{fig:musicv2}
\end{figure*}

\section{Additional Discussion of Music-guided image editing.}
\label{sec:appendix-music}

In this section, we discuss two novel use cases of \ours~ in music-guided image editing. These examples are also included in the supplementary video with accompanying audio. It is challenging to evaluate such capabilities because even humans can associate music and visual elements in various ways. Hence, we focus on qualitative results and show that \ours~ can reasonably associate music with visual elements. Specifically, we aim to demonstrate that: 1) \ours~ associates music inputs with visual edits in a consistent manner, rather than performing random edits on the input image; and 2) such associations make sense to humans.

\subsection{Music Inspired Design and Music-guided Image Variation.} We provide additional qualitative results of music-inspired design and music-guided image variations.  in \cref{fig:musicv3}. For these results, we prompted \ours~ with structured templates such as "Please modify the design of [image] based on [music]" and "modify [image] to convey the feeling of [music]". We use the same music prompt as in \cref{fig:music} from the main paper.  \ours~ was able to grasp the mood conveyed by different pieces of music and generate appropriate images. For example, it consistently associates \textit{Guren}, a Japanese Rock Song used in the anime \textit{Naruto} as its opening, with vibrant, highly-saturated, neon-light-like colors. This suits the fast-paced, highly energetic music well.  Similarly, it consistently associates \textit{Rain of Castamere}, a piece from the TV series \textit{Game of Thrones}, with a cold, lifeless atmosphere. In \cref{fig:music} from the main paper, it turns the background of a Jedi warrior to into a snowfield. In \cref{fig:musicv3}, it changes lush trees into lifeless branches. These edits align with the slow-moving, somber theme of the piece.

\subsection{Live Performance Visuals}

As a proof of concept, we used \ours~ to generate a set of visuals corresponding to an improvised performance involving four pieces. In this setup, we generated an initial image and performed image editing via structured templates. The model was prompted with 10 seconds of each piece. To further understand the reasoning behind each edit, we also prompted the model to generate captions using the template "please describe [music]." This was possible because the music captioning task is included in the pretraining dataset. Using captions, we observed that \ours~ was able to identify the tempo and associated emotions of the music. We used an iPhone to record the performance, which led to some artifacts in the recording, and this was reflected in the generated captions. Despite this, our model was still able to perform reasonable edits. For example, it associated keywords such as "soft, relaxing" with bright colors, and keywords like "suspenseful, slow" with dark tones. Notably, when the caption included "nostalgic," the model converted modern buildings into antique ones. This particular part is a segment of \textit{Departure} from the series \textit{Rurouni Kenshin}, which describes a farewell of samurais in ancient Japan. \ours~ was able to capture the essence of this piece and make appropriate edits. An audible version is included in the supplementary video.

The demo is only a proof of concept. The edits were not performed in real time. Instead, we recorded a video and retrospectively applied the edits. However, considering that we only took 10 seconds of music to prompt the model, and that the edits can be performed with structured templates without human text input, it is feasible to build a real-time system for this application using our model.

\subsection{Additional Discussions}
\begin{figure}[t]
    \centering
    \includegraphics[width=0.9\linewidth]{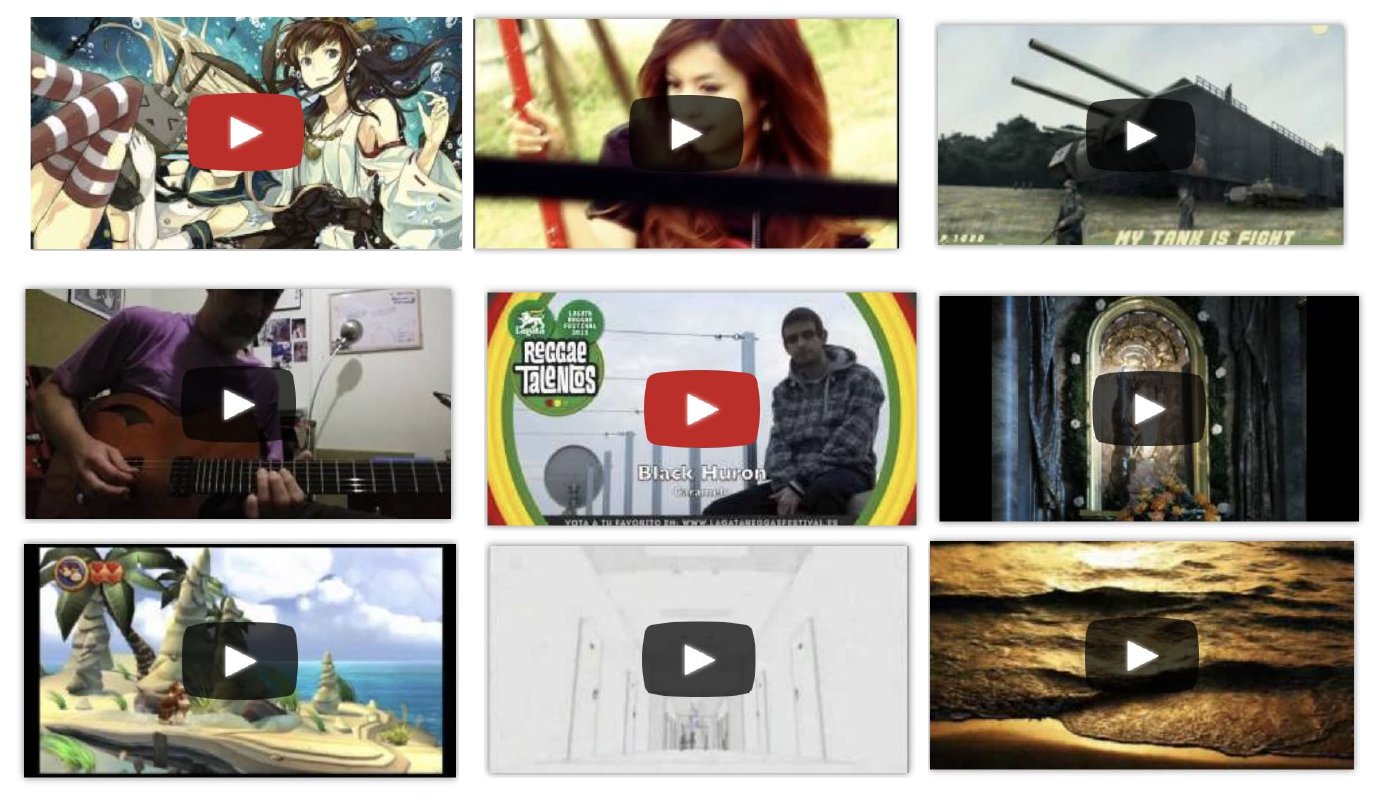}
    \caption{\textbf{Example Thumbnails of Videos with Music.} We show samples of thumbnails from the music category of the AudioSet dataset, which incorporates diverse music-image correspondences.}
    \label{fig:music_data}
\end{figure}

\textbf{How did the model learn music-visual correspondence?} There are two sources of music-visual correspondence in the training data. First, during pretraining, we incorporated an audio-guided image generation task with the prompt "Generate an image from [sound]," where the sound-image pairs came from audiovisual datasets such as AudioSet. Empirically, these pairs were curated by extracting the audio and corresponding frames from YouTube videos. When the audio piece is music, the image would be what human creators deemed suitable for that music. We show examples of images corresponding to music pieces in \cref{fig:music_data} from the AudioSet dataset. They include a diverse set of images such as album cover art, documentary stills, music videos, video game footage, and nature scenery shots. These reflect the diverse ways humans associate music with images and allow \ours~ to learn a general music-visual correspondence. Second, during instruction fine-tuning, the MM-Inst dataset includes a music-guided image editing task generated by LLM. An example would be "fit the [image] to [a peaceful, slow-moving, piano music]."

We observed that removing music-image pairs from the pretraining data led to catastrophic failures, even after fine-tuning on MM-Inst. This suggests that large-scale pretraining with natural music-visual correspondences created by humans is necessary. Interestingly, removing audio-image pairs from the pretraining data did not affect the performance as much. For example, after removing audio-image pairs from the pretraining data, the model failed to perform tasks such as "fit the [image] to [a piece of slow-moving, piano music]." However, the model could still perform tasks like "add [sound of dog barking] to the image." We hypothesize that learning simple, explicit audio-image correspondence is easy (e.g., dog sound to dog images) and can be implicitly achieved if the audio and image embeddings are well aligned with the text embedding. However, learning abstract, implicit music-image correspondence is nontrivial and requires direct training on music-image pairs.

\textbf{How useful is this capability of music-guided editing?} While music-guided editing may seem like a niche application at first glance, we have provided various examples above, including three practical use cases: music-inspired designs, music-guided image variation, and live performance visuals. One of the remaining concerns is usability. For example, would writing a text prompt and uploading music be more inconvenient than just writing a more detailed text instruction? We argue this is not necessarily the case. For example, if a user wants to have 10 good T-shirt designs but lacks expertise in fashion, it would be hard for them to write 10 detailed prompts describing all the visual elements of the designs. However, they could simply write one template, "please modify the design [an image of a blank T-shirt] to [music]," and apply it to tens or hundreds of music tracks, then pick the 10 best results. Moreover, these selected tracks can serve as a medium to apply similar designs to other objects, such as dresses. In \cref{fig:music} of the main paper and \cref{fig:musicv3}, we see that the same track leads to consistent designs across different objects. This makes music-guided designs more appealing than alternatives (e.g., randomly generating 100 images of T-shirts and picking one).

In the case of live performance visuals, we can also apply a predefined template and change the background of the stage about 10 seconds after a new piece is played. This makes it particularly suitable for impromptu performances. Since it is impossible to know what will be played or when transitions between pieces will happen, it is impractical to create stage visuals in advance. \ours~ offers unparalleled flexibility, as it allows the staff to create stage visuals that fit the piece being played with just one click.

In a similar spirit, \ours~ can also be used in bars and restaurants with a real-time "social media wall," where customers can post photos that are displayed in real time. It would be exciting if the posted photos were automatically adjusted to suit the piece being played by an impromptu artist or just the background music of the venue. Likewise, \ours~ can be used to create video filters for short videos on YouTube or Instagram. Beyond the use cases the authors have imagined, the possibilities are limitless.

\section{Additional Comparisons with Baseline Methods.}
\label{sec:appendix-baselines}
\subsection{Text-Guided Image Editing}
\begin{figure*}
    \centering
    \includegraphics[width=1.0\linewidth]{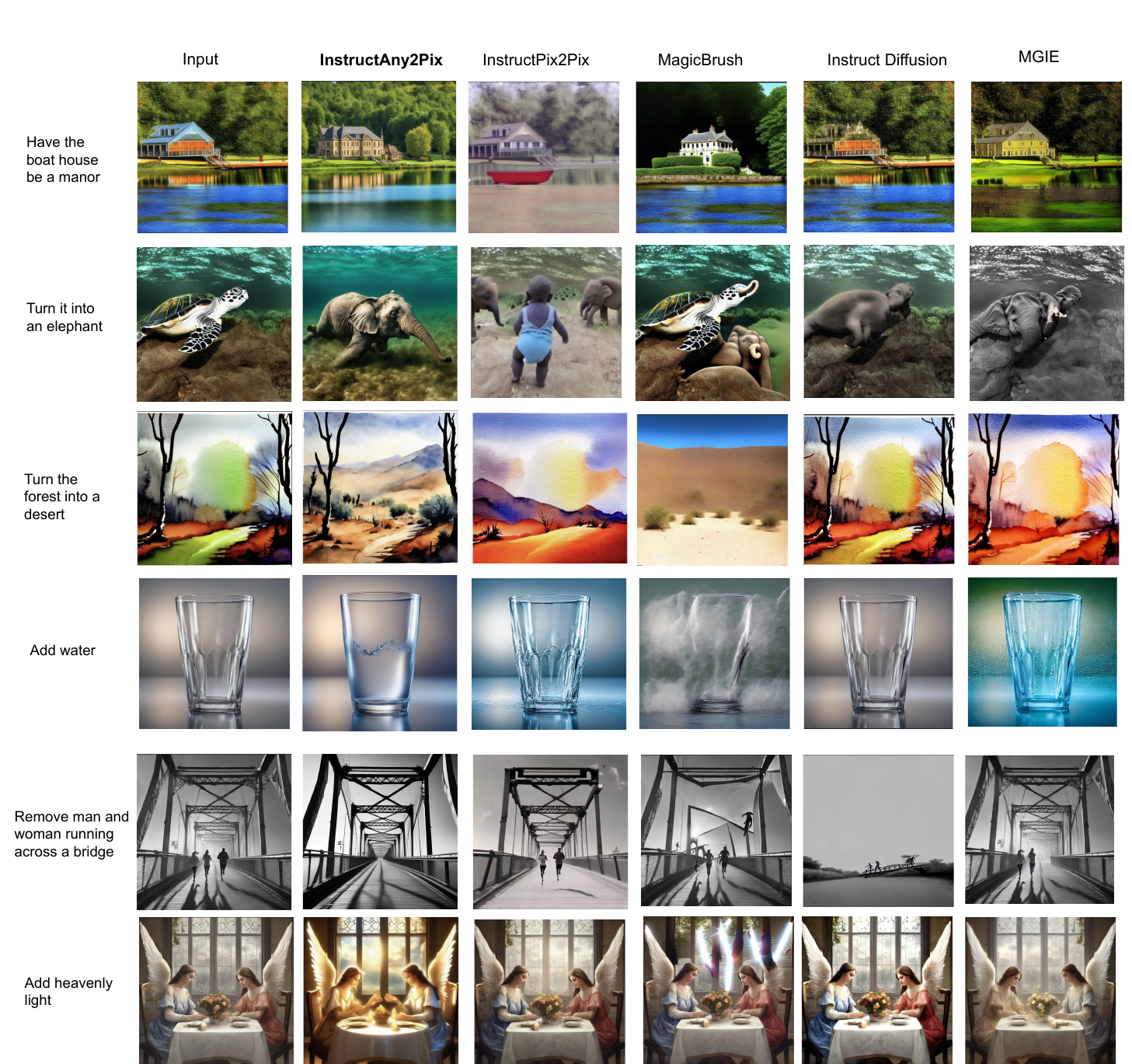}
\caption{\textbf{Qualitative Comparison Against Text-based Editing Methods.} We show editing results of different models using diverse editing instructions. The top three rows are sampled from the InstructPix2Pix dataset, and the bottom three rows are sampled from the MM-Inst-Test dataset.}

    \label{fig:text_comparison}
\end{figure*}

We provide a qualitative comparison with methods using text instructions. We compare our results against InstructPix2Pix \cite{brooks2023instructpix2pix}, MagicBrush \cite{Zhang2023MagicBrush}, Instruct Diffusion \cite{Geng23instructdiff}  and MGIE \cite{fu2023guiding} in \cref{fig:text_comparison}. For fairness, we incorporate instructions from both the InstructPix2Pix dataset and the MM-Inst-Test dataset. For MM-Inst-Test, we convert the multimodal instructions to text by replacing the multimodal token with captions of the referred image and audio. For \ours, we use the checkpoint that is not trained on InstructPix2Pix dataset for these results. 

\ours~ shows better editing results on both datasets. Particularly, on some tricky examples, such as changing a sea turtle into an elephant and adding water to the glass, \ours is the only model that can successful perform the edits. These results are exceptionally impressive, considering that \ours~ is not trained on the InstructPix2Pix dataset, unlike other methods.

\subsection{Image-Guided Synthesis}

\begin{figure*}
    \centering
    \includegraphics[width=1.0\linewidth]{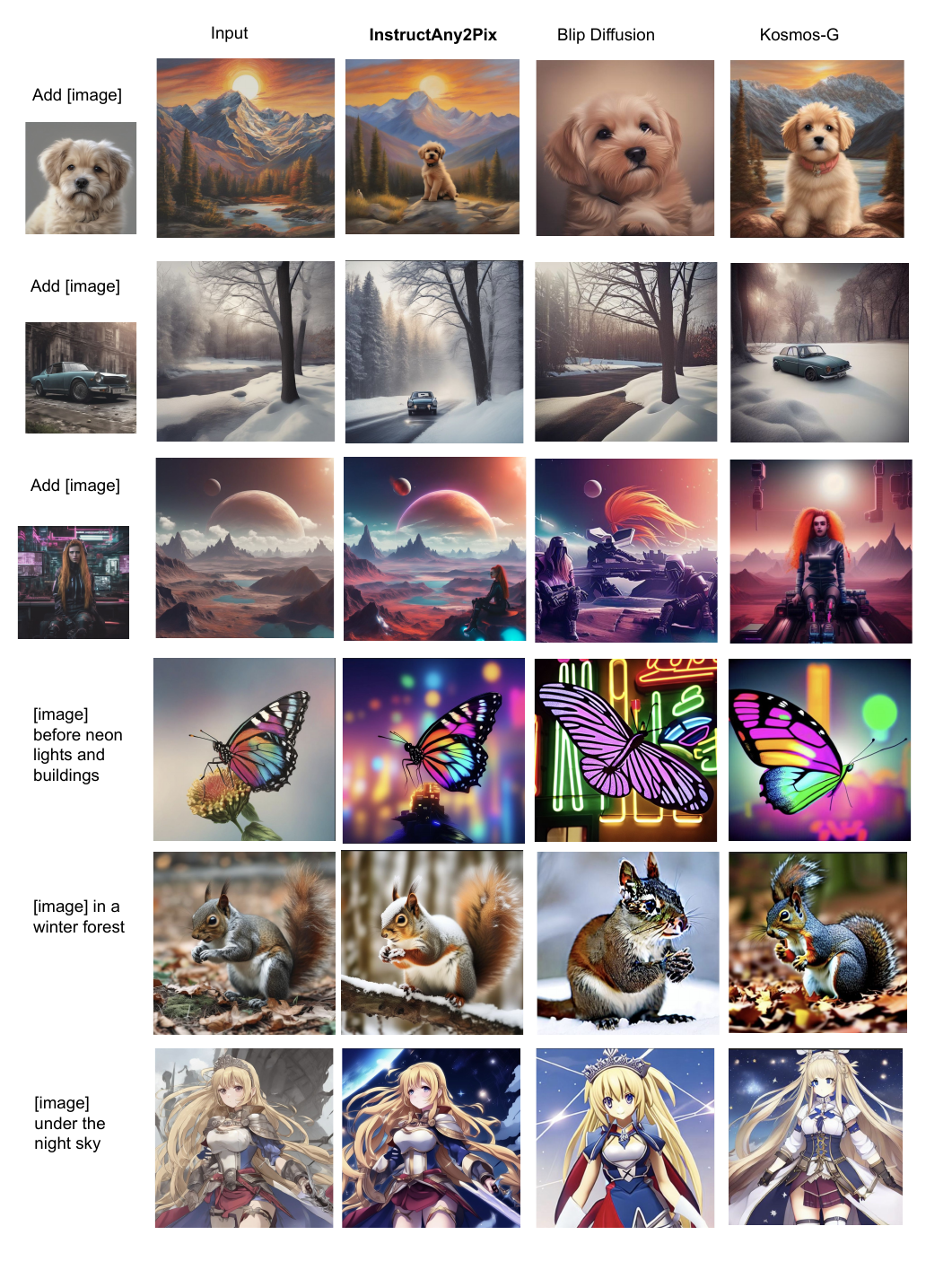}
\caption{\textbf{Qualitative Comparison Against Image-based Generation Methods.} We present generation results of different models under both single-image and multi-image setups. We employ multi-modal prompts from the Dreambooth++ dataset.}

    \label{fig:img_comparison}
\end{figure*}

We provide a qualitative comparison with multi-modal generation methods that use reference images as prompts. We compare our results against BLIP-Diffusion \cite{mangalamReversibleVisionTransformers2022} and Kosmos-G \cite{kosmos-g} in \cref{fig:img_comparison}. Visual results show that \ours~ outperforms these two baselines both in terms of generation quality and image consistency. We also conducted human evaluation using Amazon Mechanical Turk. We asked users to pick the best result in a side-by-side comparison of \ours~ and baseline methods using generations of Dreambooth++ dataset. We achieved a win rate of 79.0\% against BLIP-Diffusion and 86.2\% against Kosmos-G.

\begin{figure*}
    \centering
    \includegraphics[width=1.0\linewidth]{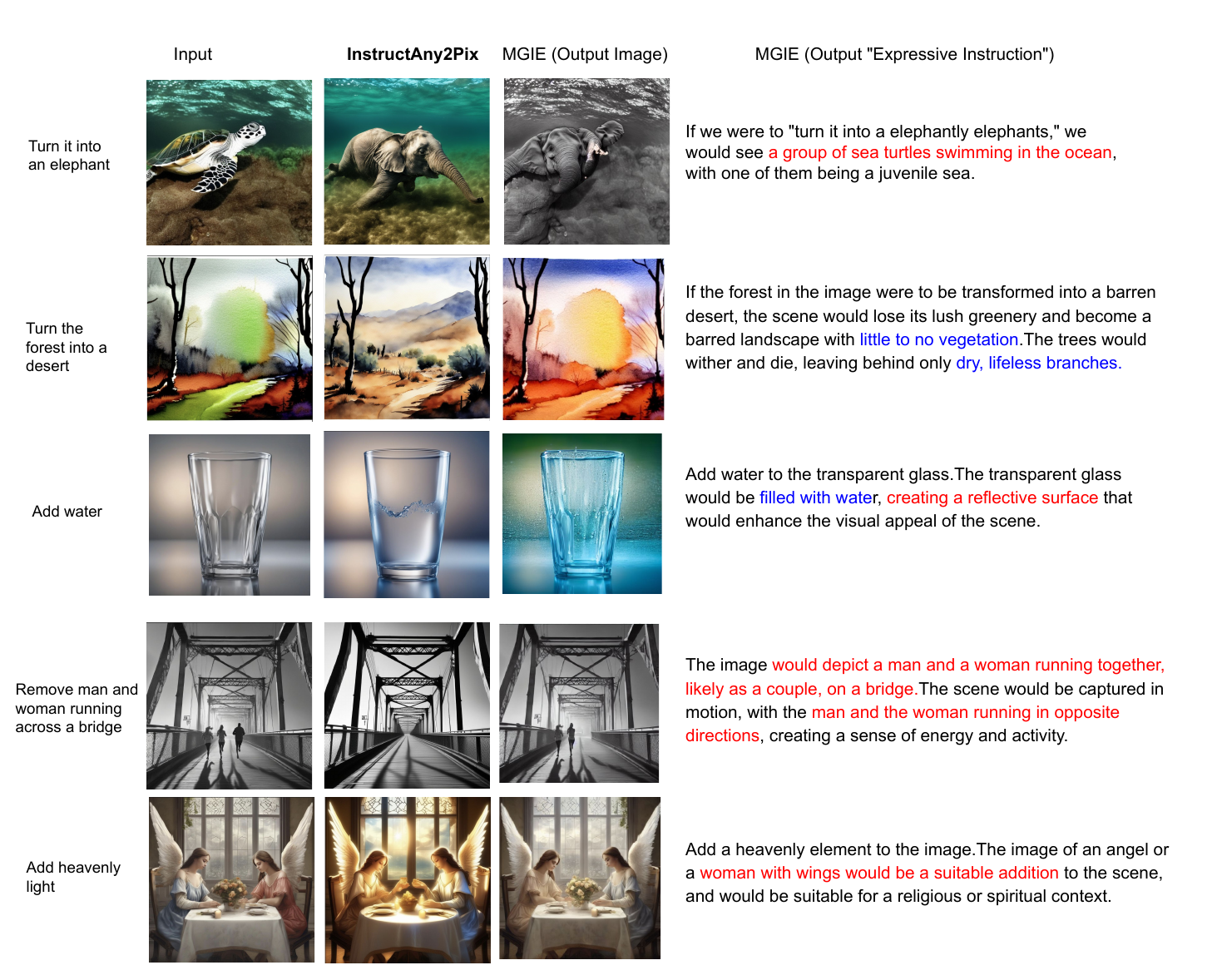}
\caption{\textbf{Analysis of failures of MGIE.} We show "expressive instructions" generated by MGIE alongside image outputs. These samples reveal the underlying reasoning process of the MGIE. We observe two failure modes. For simple instructions, MGIE can generate the correct reasoning based on instructions, but failed to apply these edits to the image. We mark these examples as \textcolor{blue}{blue}.  For more complex or abstract edits, MGIE fails to understand the instructions and generates wrong reasoning. We mark these examples as \textcolor{red}{red}. By contruct, \ours~ consistent perform the intended edits successfully. }

    \label{fig:mgie_comp}
\end{figure*}

\begin{figure*}
    \centering
    \includegraphics[width=1.0\linewidth]{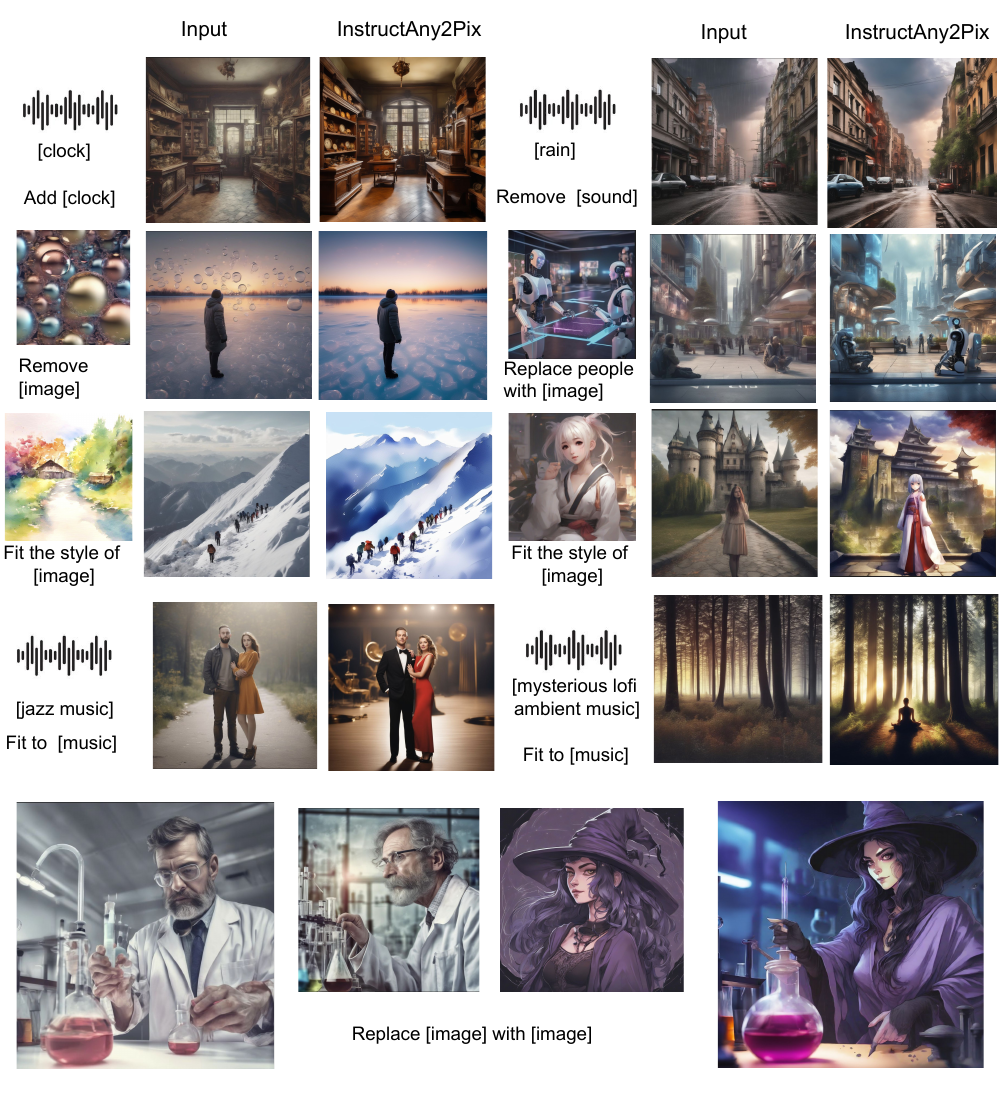}
    \caption{\textbf{Additional Results of Multi-Modal Editing.} We showcase qualitative results of multi-modal editing. \ours~ can handle a diverse set of instructions involving multiple modalities.}
    \label{fig:diverse_instruction}
\end{figure*}

\label{sec:appendix-image-baseline}
\subsection{Comparison with MGIE}

We attempted to compare against MGIE \cite{fu2023guiding}, another image editing method based on Multi-Modal Language Model, on text-based editing. However, we are unable to reproduce the results using the official checkpoint in the Github repo. As our best efforts, we provide qualitative comparisons in \cref{fig:multiturn} and \cref{fig:text_comparison} using the online inference demo hosted by Apple. We use the default parameters provided by the website for these results.  \ours~ outperforms MGIE on both complex instructions involving multiple objects, such as "Remove a man and woman running across a bridge", as well as simple ones such as "turn the forest into a desert".  

We further analyze the failure cases of MGIE in \cref{fig:mgie_comp} by making use of the "expressive instruction" generated by the online demo. These outputs reveal the underlying reasoning process of the MGIE. We observe two failure modes. For simple instructions, the model can generate the correct reasoning based on instructions, but failed to apply these edits to the image. For example, for the instruction "turn the forest into a desert", the text output of MGIE identifies relevant concepts: 
"little to no vegetation", "dry, lifeless branches". However, the generated image fails to respect these concepts. For more complex instructions such as "remove man and woman running across a bridge", the model fails to understand the intent, and believes the output image should "depict a man and a woman running together, likely as a couple, on a bridge", which is contrary to the instruction. 

This limitation can be caused by a variety of reasons: MGIE only uses InstructPix2Pix as its pretraining dataset, whose edit instructions are less diverse than those in MM-Inst. MGIE uses a MLLM to model the \textit{expressive edit instructions}, which contains \textit{what should be done} to achieve desired edits. By contrast,  \ours~ directly models the semantics of the intended output image, which contains \textit{what the output should look like}. The objective of \ours~ is more straightforward. Lastly, MLLM freezes the language model itself and only trains its adaptors and edit-heads, which may limit its capabilities.

Additionally, we note that while MGIE also make use of a Multi-Modal Language Model, they do not support "multi-modal editing" in that they only accept text-only instructions. The vision encoder of MGIE is used to only process the input image, instead of multi-modal prompts like \ours. In summary, \ours~ outperforms MGIE on text-based edits, particularly in the presence of challenging edit prompts. \ours~ also supports more flexible instructions and multi-modal prompts, making it more preferable in most practical use cases. 

\subsection{Image Editing with Multi-Modal Instruction}
We present additional qualitative results of multi-modal editing in \cref{fig:diverse_instruction}. The results demonstrate that \ours~ can effectively handle a diverse range of instructions involving multiple modal inputs.

\section{Ablation Studies}
\subsection{Pretraining}
We experiment with three pretraining setups: no-pretraining, captioning tasks only (x-to-text), and full pretraining (x-to-text and x-to-image). We report quantitative results of image-editing task on MM-Inst-Test dataset in \cref{tab:pipeline_ablation}. Full pretraining is required to achieve optimal performance.  

\begin{table}[h]
    \centering
    \caption{\textbf{Ablation Study on Pretraining Strategies.} We experiment with three pretraining setups: no-pretraining, captioning tasks only (x-to-text), and full pretraining (x-to-text and x-to-image). We report results on MM-Inst-Test dataset.  }
   \begin{tabular}{c ccc}
    \toprule
         & C$_{dir}$ & C$_{im}$ & C$_{out}$ \\
         \hline
     No Pretraining    & .071  & .795 & .207 \\
      Caption Only   & .090  & .802 & .251 \\
        Full      & \textbf{.099}  &\textbf{ .816} & \textbf{.260}\\
        \bottomrule
    \end{tabular}
    
    \label{tab:pipeline_ablation}
\end{table}

\subsection{Factors affecting Image Consistency }

\begin{figure}[h]
    
        \centering
        \includegraphics[width=1.0\columnwidth]{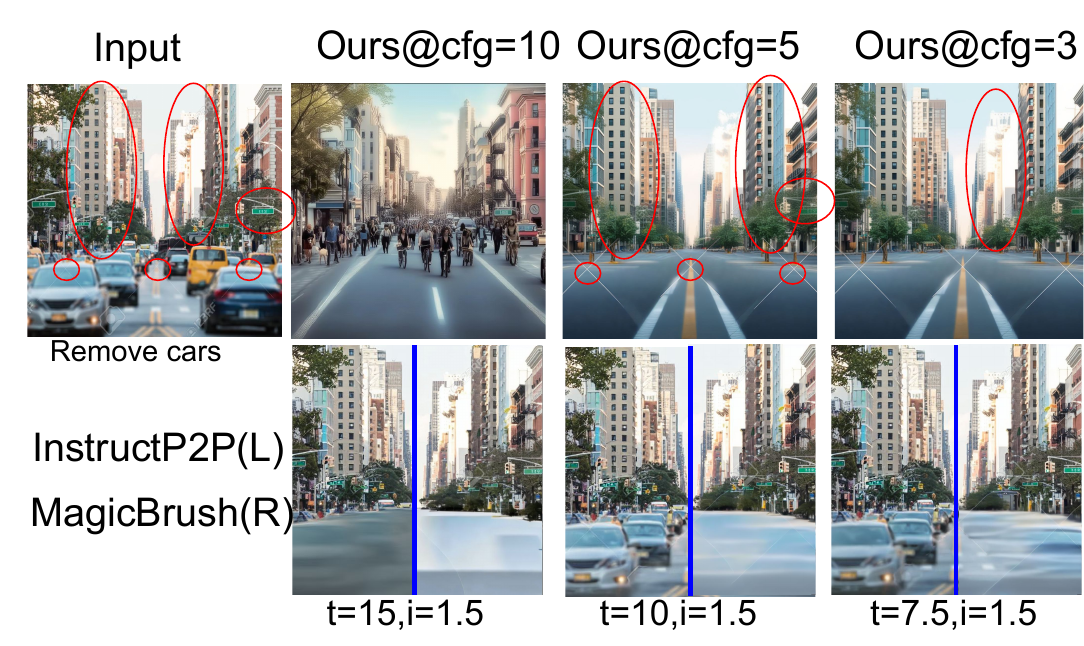}
        \caption{\textbf{Visual examples of \ours and baseline method under different classifier free guidance (CFG).} We compare \ours~ with InstructPix2Pix. InstructPix2Pix has two independent CFG for text and image. We abbreviate this as "t" and "i". Notably, our method generates artifact-free results in all setup, while other methods have visible artifact at all CFGs. }
        \label{fig:control}

\end{figure}

An important goal of image editing is to ensure the edited images can reflect the intended changes while respecting the source image. There is usually a trade-off between these two goals. The most relevant hyperparameter of \ours~ is classifier free guidance (CFG). CFG determines the degree at which the text instruction affects the generation output. We visualize edit results under different CFG in \cref{fig:control}. We find that CFG=5 is a sweet spot for achieving high quality edit results that follows the instructions while respecting the original image.

\begin{figure}[h]
    \centering
    \includegraphics[width=1.0\linewidth]{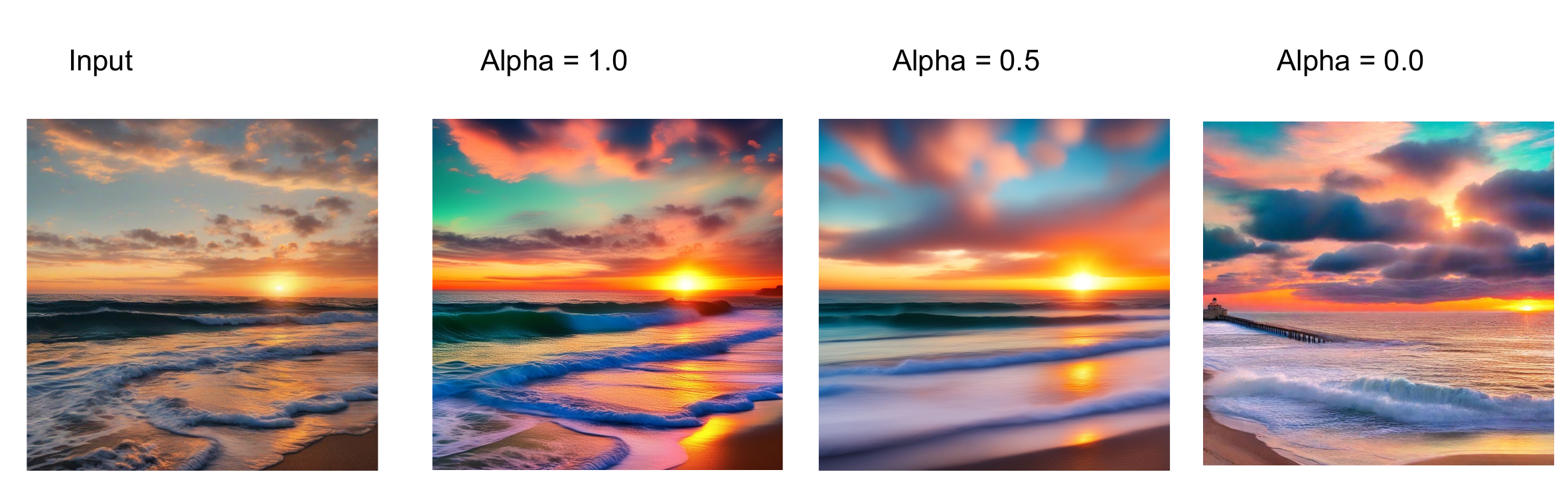}
    \caption{\textbf{Results of Varying $\alpha$ in the generation process.} We visualize the image generation conditioned on the CLIP embedding of the source image. As $\alpha$ decreases, the generation become less consistent with the source image.}
    \label{fig:alpha_appendix}
\end{figure}

In addition to CFG, we can control how well the model respects the input image by adding Gaussian to the input image in the latent space. The variance of added noise is proportional to $(1-\alpha)$ where $\alpha$ is a hyperparameter between 0 and 1. Intuitively,  when $\alpha$ is 1, there is no corruption to the image latent. When $\alpha$ is zero, the diffusion model mostly ignores the input image. We visualize this effect in \cref{fig:alpha_appendix}. For a typical use case, there is no need to corrupt the input image. We suggest setting $\alpha$ to 1.0.

\begin{figure}[h] 

 \centering
 \includegraphics[width=1.0\columnwidth]{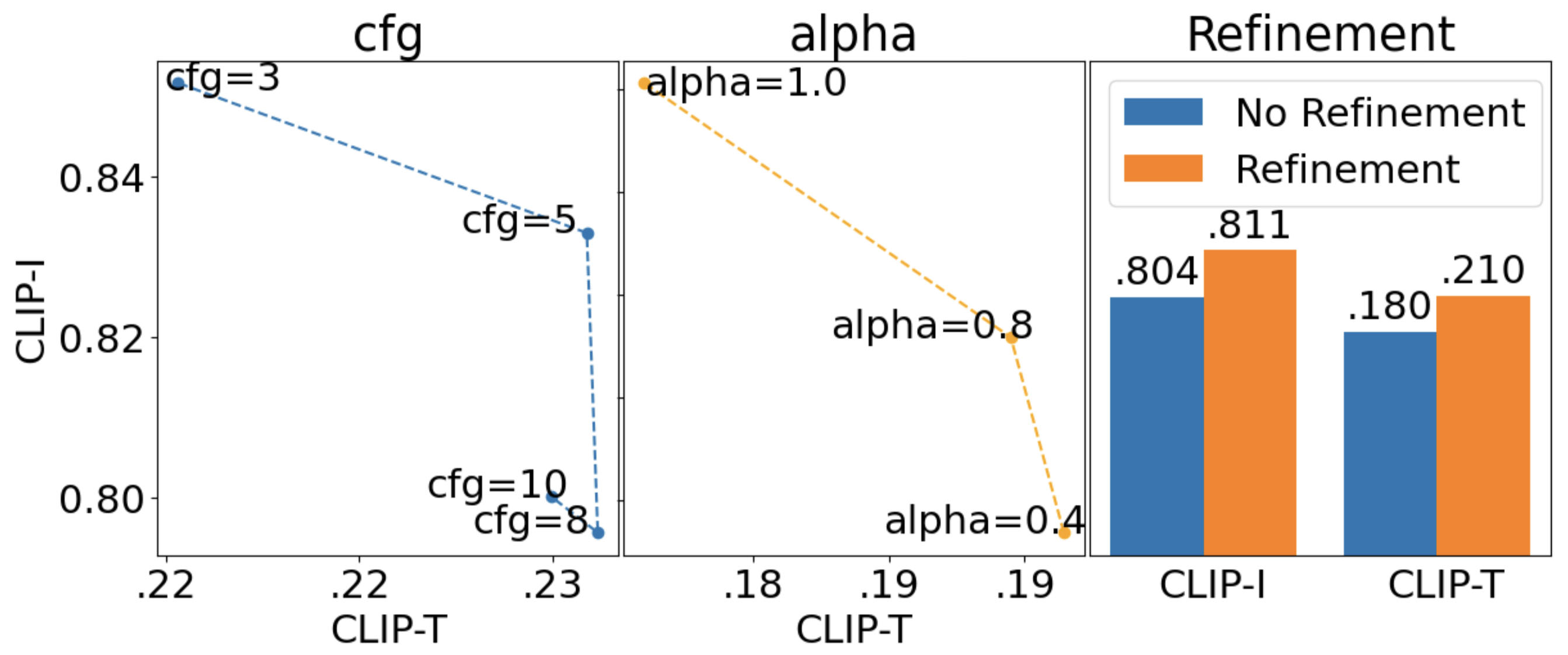}

        \caption{\textbf{How CFG, $\alpha$ and the refinement module affects instruction alignment and image consistency.} Increasing the CFG and decreasing the alpha will increase CLIP$_{out}$ (CLIP-T) and decrease CLIP$_{im}$ (CLIP-I). Adding refinement module improves both metircs. }
        \label{fig:ablation}
\end{figure}

We qualitatively evaluate the effect of these two hyperparameters on a subset of MM-Inst-Test dataset by sweeping over different values of CFG and $\alpha$. We report CLIP$_{out}$, which measures alignment with edit instructions and CLIP$_{im}$, which measures consistency with input images. We show these results in \cref{fig:ablation}.  In general, increasing the CFG and decreasing the alpha will increase CLIP$_{out}$ and decrease CLIP$_{im}$, giving the user the flexibility to balance the instruction alignment and image consistency. We also found that removing the refinement module leads to a small drop in both metrics, highlighting its effectiveness.

\subsection{Factors affecting Image Quality}

\begin{figure}
    \centering
    \includegraphics[width=1.0\linewidth]{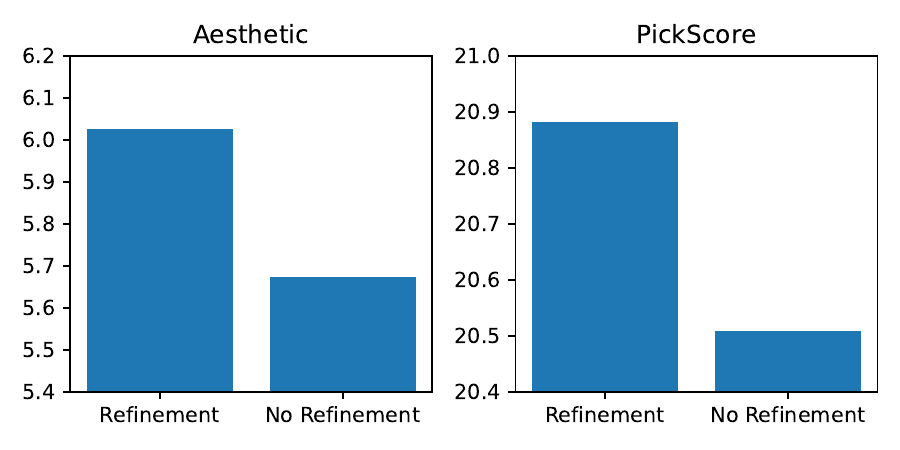}
    \caption{\textbf{ How refinement module affects image quality.} We report average Aesthetic score and PickScore \cite{Kirstain2023PickaPicAO} on 500 randomly sampled captions. Aesthetic scores only consider the image quality, while PickScore additionally takes the prompt into account.}
    \label{fig:ablation_of_refinement_appendix}
\end{figure}

The refinement module is introduced as a regularization to mitigate the effect of low quality images in the data. We sample 500 generated images and evaluate the LAION Aesthetic score and PickScore \cite{Kirstain2023PickaPicAO}, which measures image quality. Aesthetic score only considers the image quality, while PickScore additionally takes the prompt into account. We compare generations with refinement module to those without refinement module and report the results in \cref{fig:ablation_of_refinement_appendix}.  On both metrics, adding the refinement module leads to considerable improvements.  

\section{Technical Details}
\subsection{Model Architecture}

\subsubsection{Multi-Modal Encoder}
We use ImageBind \cite{girdhar2023imagebind} as our multi-modal encoder. Particularly, ImageBind uses CLIP-ViT-L as its text and image encoder. It includes an additional audio encoder that is aligned to the representation space of CLIP-ViT-L. We use the pooled token as our multi-modal embedding. It is kept frozen throughout all training stages.

\subsubsection{Diffusion Model}
We use the SDXL \cite{podell2023sdxl} as our diffusion model. It was originally conditioned on CLIP-ViT-G and CLIP-ViT-L text features. We incorporate an MLP projection layer following \cite{ye2023ip} that maps the ImageBind embedding to the dimension of cross-attention layers. During the pretraining process, the loss is 

\begin{equation}
\mathcal{L}_{\text{diff}} = \mathbb{E}_{t} \left[ \lVert \epsilon_{t} - p_{\theta}(z_t,C_{t},C_{g},C_{l}) \rVert_2^2 \right]
\label{eq:diff_loss}
\end{equation}

where $p_{\theta}$ is the U-Net, $\epsilon_t$ is the noise at timestamp $t$, $z_t$ is the noised image latent sampled in the forward diffusion process at time $t$, $C_t$ is the CLIP embedding of captions. $C_g$ is the embedding of the whole image, $C_l$ is the embedding of a cropped region. The cropped region is sampled from an object detector or a uniform distribution of bounding boxes at a 1:1 ratio. At each training step, $C_t$, $C_g$, $C_l$ are randomly dropped independently with a probability of 0.2. 

Diffusion Model is not used in the instruction fine-tuning stage.

\subsubsection{Refinement}
We adopt a decoder-only transformer with 24 layers and a hidden size of 1024. We also incorporate an MLP projector that maps ImageBind features to the hidden dimension of the transformer. Another MLP projector is used to map the output of the transformer back to the dimension of ImageBind features.

\subsubsection{Multi-Modal Large Language Model}

We use Vicuna-7B \cite{vicuna} as our base model. We made no additional changes to the LLM architecture other than adding input and output projectors, which are two two-layer MLPs. The input projector maps the embedding of multi-modal encoder to the dimension of MLLM's hidden states. The output project maps the extracted hidden states from the MLLM to the dimension of encoder embeddings. 

\subsubsection{Parameter Count}
We report the total number of parameters in each module in \cref{tab:num_count}. In total, our model has around 10B parameters.

\begin{table}[h]
\caption{\textbf{Number of Parameters in \ours}. We report the total number of parameters in each module. }
    \centering
    \begin{tabular}{c c}
    \toprule
     & Params \\ 
    \hline
       LLM  & 7B \\
       SDXL & 2.5B \\ 
       Refinement  & 71.1M \\ 
       Projectors  & 62.9M \\
    \bottomrule
    \end{tabular}
    
    \label{tab:num_count}
\end{table}

\label{sec:appendix_model_arch}
\subsection{Training Dataset}
\label{sec:appendix_training_dataset}
\subsubsection{Paired Training Data}

We use SoundNet \cite{aytar2016soundnet}, VGG-Sound \cite{chen2020vggsound}, and AudioSet \cite{gemmeke2017audio} for image-audio pairs. These datasets consist of videos with audio. We extract the audio and the middle frame from the video to create audio-image pairs. SoundNet consists of 802,724 audio-image pairs, AudioSet consists of 2 million audio-image pairs, and VGG-Sound consists of 197,958 pairs. Particularly, out of 2 million videos in AudioSet, 1 million are under the music category. These videos can be music videos, concerts, documentaries, and other kinds of videos that use music as background, such as news programs.

We also make use of audio captions from MusicCaps \cite{agostinelli2023musiclm} and AudioCaps \cite{audiocaps} to create text-audio pairs. These two datasets provide text captions for subsets of AudioSet. They do not introduce new audio files. We use LAION-Aesthetic-3M \cite{schuhmann2022laion} for text-image alignment, which consists of 2,209,745 valid image URLs at the time of data fetching (Sep 2023). All these datasets are used in the alignment process.

\subsubsection{Instruction Tuning Dataset (MM-Inst)}
\begin{figure*}[t!]
    \centering
    \includegraphics[width=1.0\linewidth]{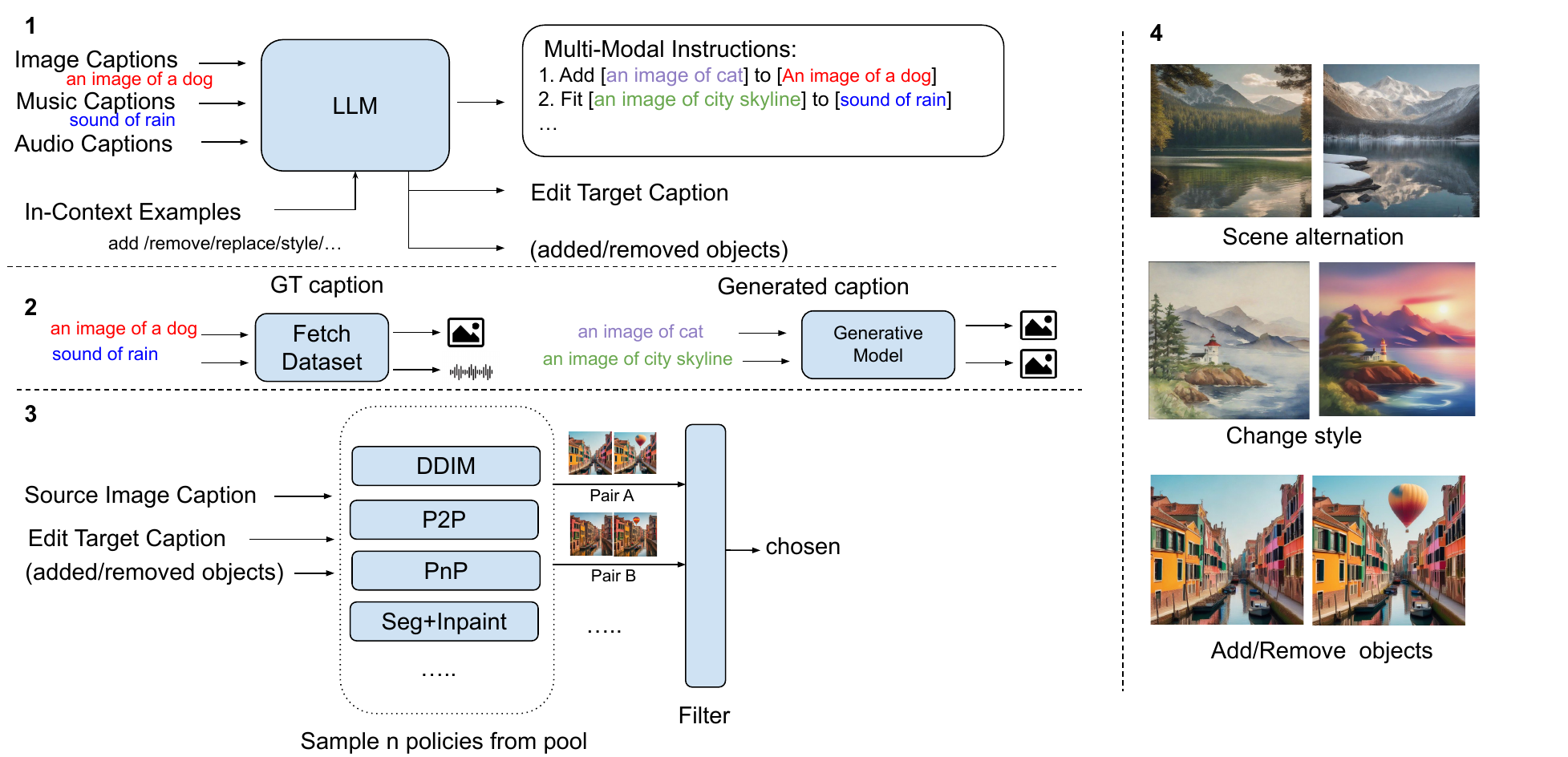}
    \caption{Data generation Pipeline. 1. We prompt LLM with sampled captions and examples to generate a diverse set of instructions. 2. We obtain reference music, audio, and images by either generating them using SDXL and AudioLDM2. If the caption corresponding to a ground truth caption, we directly fetch the corresponding media. 3. We employ a mixture of methods to generate candidate image pairs for each instruction and filter them using CLIP score. 4. We show some example image pairs in the filtered dataset on the right.}
    \label{fig:appendix-dataset-explainer}
\end{figure*}

As described in \cref{sec:main-mm-inst-data} of the main paper,  MM-Inst was generated in three steps. We show this in \cref{fig:appendix-dataset-explainer}.

In the first step, we prompt a large language model (Llama 2) to  generate creative instructions. Each instruction contains the caption of the input image, the caption of the output image, the text instruction and optionally captions of reference images and audio. We also ask the model to explicitly mark objects that need to be added to the scene. This information is used later in stage 3 to generate input-output image pairs.

Example instructions include adding, dropping, removing, or replacing objects as well as other free-form instructions.  To generate a diverse set of image editing instructions, we prompt the language model to create editing instructions based on captions of sampled LAION images. We use BLIP-2\cite{li2023blip} to generate these captions instead of using the captions provided in the dataset, because the provided captions are noisy alt text that is not natural English.  To further increase the diversity of audio-related instructions, we further prompt the language model to generate instructions involving ground-truth music captions from LP-MusicCaps and AudioCaps. We observe that without this step, the language model tends to only generate simple audio captions such as "sound of water" or "sound of rain" and fails to incorporate complex descriptions of music. In total, we generate 500k instructions. 

In the second step, we collect or generate reference images and audios using captions created in the first step. There are two types of captions. The first type appears in instructions generated by explicitly prompting the LLM with ground truth captions of music, sound, and images. For these captions, the corresponding media can be directly used. The second type is generated solely by the LLM. We use AudioLDM2 \cite{liu2023audioldm} to generate audio and music, and use SDXL to generate images. We generate 5 samples for each caption and use CLAP \cite{laionclap2023} and CLIP \cite{radford2021learning} to find the samples that align best with the caption. 

In the last step, we employ a diverse set of methods mentioned in \cref{sec:main-mm-inst-data} to create input-output image pairs.

Compared with the InstructPix2Pix dataset, MM-Inst has the following advantages. First, it uses BLIP-generated captions which are grounded with real images instead of the raw caption from LAION, which can be very noisy. Second, it uses a variety of techniques to generate paired data instead of solely relying on Prompt2Prompt. In particular, we observe that the segmentation+inpainting pipeline generates high-quality results for object removal when the segmentation model can correctly localize the target object. Third, we filter the data using both CLIP and Aesthetic score and consider both prompt alignment and generation quality. In contrast, InstructPix2Pix only uses CLIP score as the filtering mechanism. Lastly, MM-Inst incorporates multi-modal inputs, which makes it uniquely suitable for our multi-modal editing tasks. We showcase these differences in \cref{tab:dataset_comp_1}. We provide examples from both datasets in \cref{tab:examples_dataset}.

\subsection{Evaluation Dataset}

\subsubsection{MM-Inst-Test}

To generate a diverse set of prompts, we ask the MTurk workers to generate creative edit instructions using different captions sampled from LAION. We also require the MTurk workers to generate different types of edits for the same caption. We do not generate ground truth target images. For reference images and audio pieces used in the instructions, we use SDXL \cite{podell2023sdxl} to generate images and AudioLDM2 to generate \cite{liu2023audioldm2} audio. Results are filtered by CLIP\cite{radford2021learning} score for images and CLAP\cite{laionclap2023} score for audios. 

\begin{figure}
    \centering
    \includegraphics[width=1.0\linewidth]{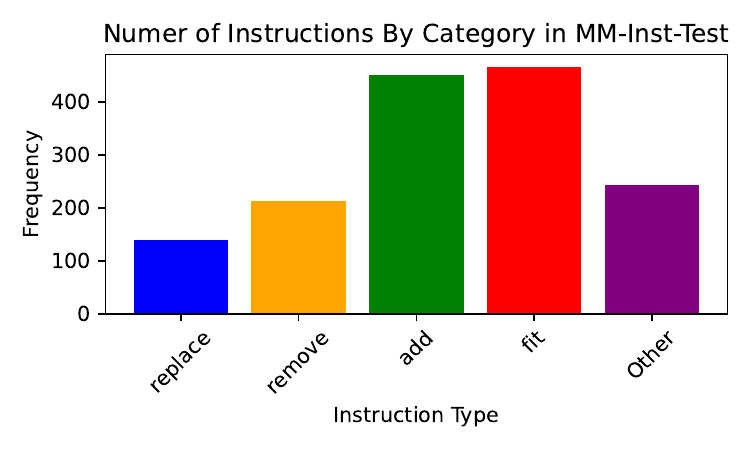}
    \caption{\textbf{Distribution of Instructions in MM-Inst-Test dataset.} We show the number of instructions. "Fit" refers to combining music and image, fitting an image to the style of another image, or other organic ways of combining different modalities together. "Others" include all instructions that cannot be classified as other categories, such as "transform [image] into a night scene with the sound of [sound]".}
    \label{fig:test_hist}
\end{figure}

One of the challenges in generating the test dataset is that there are instances of bad format and low quality. We prompt each MTurk worker to generate five different edit prompts for each caption. After a batch is collected, we manually identify the problematic instructions and redistribute them to a new set of workers. In order to reach 1,500 valid instructions, we collect a total of 1795 instructions. We show the distribution of different types of instructions in \cref{fig:test_hist}.

\clearpage
\subsubsection{Dreambooth++}
Dreambooth is commonly used to evaluate generative models that support image prompts. However, its evaluation protocol consists of only a single reference image and a text prompt. It is not suitable to evaluate models that can take multiple reference images, such as Kosmos-G \cite{kosmos-g} and \ours. Moreover, the classes have limited diversity and only have two live subjects (cats and dogs). These two live subjects accounts for 9 out of 30 subjects. All other objects are small, still objects, such as backpacks. There are no medium-to-large objects such as trees or bicycles.  To provide a diverse and fair comparison, we propose Dreambooth++, which contains 30 subjects and 30 prompts. In total, there are 900 combinations. We use SDXL to generate all subject images. For each prompt, we also generate a background.

We conduct evaluation on two benchmarks: The single-image setup is similar to Dreambooth, which requires generating a given subject under different context prompts. The multi-image task requires generating a new image by combining a subject image and a background image. For the multi-image task, we use a segmentation model to localize the edited objects in the new image. We report DINO similarity of the cropped subject with the reference subject image as DINO$_{sub}$ and the DINO similarity of the generated image with the reference background image as DINO$_{ref}$. We also report the recall rate of the segmentation model. These results are listed in the main paper \cref{tab:exp2}.

We show the full list of subjects in \cref{tab:db_dataset_subjects} and compare it with Dreambooth. DreamBooth++ offers a more diverse range of subjects ranging from animals, humans, large structures, and small items.

% \clearpage
\label{sec:appendis_evaluation_dataset}
\subsection{Compute}
\label{sec:training_details}
We use AdamW optimizer with a learning rate of 1e-6. We use 8 Nvidia A6000 GPU for our experiments. We train the model for 2 epoch, which takes around two days. The diffusion model is trained on 8 A5000 GPU with AdamW optimizer, a learning rate of 1e-6 for 4 days (40000 steps).

\begin{table*}
    \centering
    \caption{Comparison of MM-Inst dataset and Instruct Pix2Pix dataset.}
    \begin{tabular}{c c c c}
    \toprule
          & MM-Inst & MM-Inst-Test & InstructPix2Pix \\
         \hline 
  Caption  Source & LAION-Aesthetics & LAION-Aesthetics & LAION-Aesthetics \\
  Input Caption  Generation & BLIP2 & BLIP2 & Noisy WebData \\
  \hline
  Instruction Generation & Llama 2 & Human & GPT-3 \\ 
  \hline
  \multirow{4}{*}{Paired Data Generation} & DDIM & - &  \multirow{4}{*}{Prompt2Prompt} \\ 
  & Prompt2Prompt & - &  \\
  & Plug-and-Play  & - &  \\
  & Segmentation+Inpaint  & - &  \\
  \hline
  Filtering & Multiple Metrics & Human & CLIP \\
  \hline
  Modality & Image,Text,Audio & Image,Text,Audio & Text \\ 
  \hline
  Size &500,000 & 1,500 & 313,010 \\ 
  
    \bottomrule
    \end{tabular}
    
    \label{tab:dataset_comp_1}
\end{table*}

\begin{table*}
    \centering
    \caption{\textbf{Examples of Instructions from MM-Inst dataset and Instruct Pix2Pix dataset.} Captions are marked by [.]. MM-Inst offers a better set of captions and instructions. }
    \begin{tabular}{p{6cm} p{6cm}}
    \multicolumn{2}{c}{MM-Inst}\\
    \hline
    Instruction &  Output  \\
         \hline 
          Please incorporate [an image of cannon fire] into [an image of a pirate ship sailing on the high sea] & An image of a pirate ship firing at a British Navy warship, fire burning on the ship  \\
         \hline 
          Remove [sound of car accelerating] from [an image of people driving in the countryside road] & An image of a quiet countryside road \\
         \hline
         Replace [sound of dog barking] with [sound of a cute cat] for [an image of a dog at the beach] & An image of a cat at the beach \\
         \hline
       Change [an image of a woman wearing sunglasses in Paris] to the style of [an image of a Renaissance painting of a noble lady] & A Renaissance painting of a woman wearing sunglasses in Paris \\
         \hline
         Make [an image of a cute girl in a school uniform] fit the atmosphere of [a piece of music of stellar constellations] & An image of a cute girl in a school uniform under the night sky \\
         \\
         \hline
             \multicolumn{2}{c}{InstructPix2Pix}\\
             \hline
 [misurina XIII... by roblfc1892] have it be a stamp & Stamp... misurina XIII... by roblfc1892 \\
    \hline
  [Manarola during sunrise - Cinque Terre]  it is foggy & 
         Manarola during foggy sunrise - Cinque Terre
    \end{tabular}
    
    \label{tab:examples_dataset}
\end{table*}

\begin{table*}[h]
\centering
\caption{\textbf{Subjects of DreamBooth++ and DreamBooth Dataset.} DreamBooth++ offers a more diverse range of subjects ranging from animals, humans, large structures, and small items.}

\begin{tabular}{c c}
\hline
\toprule
\textbf{DreamBooth++ Dataset Subjects} & \textbf{DreamBooth Subjects} \\ \hline
a cute cat & backpack \\ \hline
a cute dog & backpack\_dog \\ \hline
a colorful butterfly & bear\_plushie \\ \hline
a colorful bird flying low over a body of water & berry\_bowl \\ \hline
spotted horse & can \\ \hline
an image of a squirrel & candle \\ \hline
an african elephant walking through a grassy field & cat \\ \hline
an image of cute anime girl & cat2 \\ \hline
\hline
an anime princess with long blonde hair and two swords & clock \\ \hline
the woman in a black dress holding a fan & colorful\_sneaker \\ \hline
a man standing at a podium & dog \\ \hline
a cyberpunk style & dog2 \\ \hline
an image of a scientist & dog3 \\ \hline
an image of an astronaut & dog5 \\ \hline
an artistic painting of a woman with blonde hair & dog6 \\ \hline
\hline
a wooden bridge & dog7 \\ \hline
car & dog8 \\ \hline
traffic lights & duck\_toy \\ \hline
train & fancy\_boot \\ \hline
tree & grey\_sloth\_plushie \\ \hline
bicycle & monster\_toy \\ \hline
an image of a robot & pink\_sunglasses \\ \hline
\hline
tablet & poop\_emoji \\ \hline
telescope & rc\_car \\ \hline
the skull & red\_cartoon \\ \hline
vase & robot\_toy \\ \hline
wand & shiny\_sneaker \\ \hline
an image of chair & teapot \\ \hline
an image of empty glass & vase \\ \hline
a cowboy hat & wolf\_plushie \\ \hline
\bottomrule
\end{tabular}

\label{tab:db_dataset_subjects}
\end{table*}

\section{Error Bars}
\label{sec:error_bars}
We report the margin of error of 95\% confidence interval of main results in \cref{tab:err_1} and \cref{tab:err_2}.
\begin{table*}[h!]

    \centering
    \caption{Multi-Modal Image Editing on MM-Inst-Test Dataset and Text-based Image Editing on InstructP2P Dataset.The best number is \textbf{bolded} and second-best is \underline{underlined}. }
    \begin{tabular}{c cccHH cccHHH}
    \toprule
         & \multicolumn{5}{c}{MM-Inst} &\multicolumn{5}{c}{InstructP2P}  \\
         \cmidrule(r){2-6} \cmidrule(r){7-11}
         & CLIP$_{dir}$ & CLIP$_{im}$ & CLIP$_{out}$ & DINO. & Win.  & CLIP$_{dir}$ & CLIP$_{im}$ & CLIP$_{out}$ & Zero-shot & Win. \\

      \hline
          Ours(T)    & \textbf{.095±.003}  & \underline{.856±.001} & \textbf{.270±.002} & \textbf{-} & - & \underline{.147±.003} & \underline{.808±.003} & \textbf{.312±.002} & \checkmark & - \\
      InstructP2P      & \underline{.091±.003} & .824±.002 & \underline{.243±.002} & - & .712 & .145±.003 & .742±.004 & .241±.002 & \xmark & .646 \\
      MagicBrush & .084±.004 & .807±.006 & .199±.002 & - & .707 & \textbf{.165±.004} & .760±.006 & .250±.001 & \xmark & .698\\
    InstructDiff. & .066±.003 & \textbf{.940±.004} & .193±.002 & - & .746±.003 &.126±.002 & \textbf{.857±.003} & .301±.002 & \xmark & .631 \\
  %   \hline
  % Ours(T+I)    &\textbf{.215}  & .759 & \textbf{.271} & .51/.58 & - \\
  % Blip-Diff.    & .132  & \textbf{.790} & .214 & .49/.60 & .790 \\          
  %  Kosmos-G    & .198  & .656 & .270 & .60/.33 & .862 \\
    \bottomrule
    \end{tabular}
    
    \label{tab:err_1}
\end{table*}

\begin{table*}[h!]

    \centering
    \caption{Image Conditioned Generation on DreamBench++ Dataset. $C_{dir},C_{im},C_{out}$ is abbreviated form of  CLIP$_{dir}$, CLIP$_{im}$ and CLIP$_{out}$. For multi-image setup, numbers are reported in DINO$_{ref}$/DINO$_{sub}$ format. The best number is \textbf{bolded} and second-best is \underline{underlined}. }
    \begin{tabular}{c cccHH cccHH}
        \toprule
         & \multicolumn{5}{c}{Single-Image} &  \multicolumn{5}{c}{Multi-Image}\\
         \cmidrule(r){2-6} \cmidrule(r){7-11}
         & C$_{dir}$ & C$_{im}$ & C$_{out}$ & DINO. & DINO  & C$_{dir}$ & C$_{im}$ & C$_{out}$ & DINO & Recall \\
         \hline
   % \textbf{-} & \textbf{-} \\
     % \hline
          Ours(T+I)    & \textbf{.147±.004}  & \underline{.810±.004} & \textbf{.260±.002} & \textbf{-} & \textbf{.688} & \underline{.154±.004} & \textbf{.789±.004} & \textbf{.309±.002} & \textbf{.625}/\underline{.471} & \textbf{.841}\\
                BLIP-Diffusion   & .089±.005 & .779±.005 & .231±.002 & - & .660 & .091±.005 & .701±.005 & \underline{.292±.002} & \underline{.526}/.422 & .693 \\
       Kosmos-G & \underline{.126±.005} & \textbf{.843±.005} & \underline{.251±.002} & - & .683 & \textbf{.166±.005} & \underline{.740±.005} & .286±.002 & .485/\textbf{.476} & \underline{.812}\\

  %   \hline
  % Ours(T+I)    &\textbf{.215}  & .759 & \textbf{.271} & .51/.58 & - \\
  % Blip-Diff.    & .132  & \textbf{.790} & .214 & .49/.60 & .790 \\          
  %  Kosmos-G    & .198  & .656 & .270 & .60/.33 & .862 \\
    \bottomrule
    \end{tabular}
    
    \label{tab:err_2}
\end{table*}

\section{Border Impacts}
\label{sec:border_impacts}
\ours~ aims to improve the flexibility of image editing models by incorporating multi-object, multi-modal prompts. In particular, \ours~ uniquely enables a set of creative music-based applications such as music-inspired design. However, just as any other image-edit models, \ours~ can be used to for fraud and deception. Particularly, the ability to synthesize multiple images using text instructions can be used to create fake, deceptive images with high quality. Hence, it is important to employ guardrails when deploying \ours~ to end-user products. 

% particular to useful as an extra step before the  release of new T2I models that mitigates the biases without sacrificing image quality. However, it may also inadvertently perpetuate new biases, such as those against non-binary gender. Therefore, we suggest users to take extra caution when dealing with these situations. 

\section{Safe Guards}
\label{sec:safeguard}
\ours~ is based on the diffuser \cite{von-platen-etal-2022-diffusers} library. It should be used with the standard safeguards, including NSFW safety checker and hidden watermarks.

\section{License}
\label{sec:licenses}
We makes use the following models: CLIP (MIT license), PickScore(MIT license), LAION Aesthetics predictor (MIT license),  SDXL( CreativeML Open RAIL++-M License), LLAMA 2 (Llama 2 Community License Agreement),  Vicuna (Apache2 license). BLIP-2 ( BSD-3-Clause license)

We use the following dataset SoundNet (MIT license), VGG-Sound (CC BY 4.0 license), AudioSet (CC BY-SA 4.0 license), MusicCaps (CC BY-SA 4.0), AudioCaps (MIT License) LAION (MIT License).

LAION dataset is currently unlisted publically due to a safety review. 

\section{Human Instructions}
\label{sec:human_instructions}
In this section, we report the instructions used to generate human evaluation results and image captions. 

The following prompt is used to generate the caption 

\begin{tcolorbox}[colback=blue!5!white, colframe=blue!75!black, title=Instruction to Write Prompts for MM-Inst-Test Dataset, breakable]
Your goal is to generate multimodal image edit instructions including a source image caption, an edit instruction, and a target image caption. The edit instruction can involve other reference image and audios.
Some concrete examples are

1.add [image:fireworks] to [image:base:a city skyline at night] == [image:result:an image of a city skyline at night with fireworks]

2.remove [audio: water stream] from [image:base:a painting of a cabin by the lake at sunset] == [image:result:a painting of a cabin by a corn field at sunset]

3.fit [image:base:an image of empty street] to [audio: upbeat electronic music]==[image:result:an image of vibrant city street]

4.fit [image:base:a city skyline] to the style [image:an impressionist painting] == [image:result:an impressionist painting of a city skyline]

5. replace [image:people] with [image:wildlife] in [image:base:a painting of two people standing in a field surrounded by hay bales] == [image:result:a painting of wildlife in a field surrounded by hay bales]

The typical examples can be adding, removing, replacing objects, image style transfer or fitting image to audio. However, you are encouraged to be creative.
use [audio:xxx] to mark audio inputs, use [image:xxx] to mark reference image inputs. use[image:base:xx] to specify source image,use [image:result:xxx] to specify the results
use == to separate instruction and results. 

IMPORTANT : The Removed or Replaced Object should exist in the Orignal Source Image.
Special Instructions for removal/replace commands.

When removing objects, do try to remove objects not explictly mentioned in the caption

When removing objects, do NOT try to use words like xxx without yyy, simply drop the removed objects
Examples: Remove [audio:piano music] from [image:base:an image of a room with piano]

Good Example 1: [image:result:an image of an empty room]

Good Example 2: [image:result:an image of a room]

Bad Example : [image:result:an image of a room without piano]
\end{tcolorbox}

The following instruction is used to collect human feedbacks:

\begin{tcolorbox}[colback=blue!5!white, colframe=blue!75!black, title=Instruction for Human Evaluation,breakable]
Task Description

Your task is to compare the output images from two image edit models based on the following criteria:

Alignment: Does the model follow the instruction accurately?

Quality: How good is the model generation output?

Information Loss: How well does it respect the original inputs?

Note: Some models may crop the source image at the center. Please do not consider cropping as a factor in your judgment.

[images and instructions here]

Given the criteria, which of the edit output among Image A and Image B is better?

\end{tcolorbox}

\end{document}